\documentclass[11pt,a4paper,twoside]{article}
\makeatletter\if@twocolumn\PassOptionsToPackage{switch}{lineno}\else\fi\makeatother

\usepackage{amsfonts,amssymb,amsbsy,latexsym,amsmath,tabulary,graphicx,times}
\usepackage{subcaption}

\usepackage[T1]{fontenc}
\usepackage{url,multirow,morefloats,floatflt,cancel,tfrupee,multicol}

\usepackage[ruled]{algorithm2e}%
\usepackage{algorithmic}
\usepackage{hyperref}

\graphicspath{{Images/}}

\usepackage[title]{appendix}

\usepackage{comment}
\makeatletter

\AtBeginDocument{\@ifpackageloaded{textcomp}{}{\usepackage{textcomp}}}
\makeatother
\usepackage{colortbl}
\usepackage[dvipsnames]{xcolor}
\usepackage{pifont}
\usepackage[nointegrals]{wasysym}
\usepackage[superscript,biblabel]{cite}
\makeatletter

\@ifundefined{etal}{\def\etal{\textit{et~al}}}{}

\AtBeginDocument{
\expandafter\ifx\csname eqalign\endcsname\relax
\def\eqalign#1{\null\vcenter{\def\\{\cr}\openup\jot\m@th
  \ialign{\strut$\displaystyle{##}$\hfil&$\displaystyle{{}##}$\hfil
      \crcr#1\crcr}}\,}
\fi
}

\@ifundefined{tsGraphicsScaleX}{\gdef\tsGraphicsScaleX{1}}{}
\@ifundefined{tsGraphicsScaleY}{\gdef\tsGraphicsScaleY{.9}}{}
\def\checkGraphicsWidth{\ifdim\Gin@nat@width>\linewidth
	\tsGraphicsScaleX\linewidth\else\Gin@nat@width\fi}

\def\checkGraphicsHeight{\ifdim\Gin@nat@height>.9\textheight
	\tsGraphicsScaleY\textheight\else\Gin@nat@height\fi}

\def\fixFloatSize#1{}
\let\ts@includegraphics\includegraphics

\def\inlinegraphic[#1]#2{{\edef\@tempa{#1}\edef\baseline@shift{\ifx\@tempa\@empty0\else#1\fi}\edef\tempZ{\the\numexpr(\numexpr(\baseline@shift*\f@size/100))}\protect\raisebox{\tempZ pt}{\ts@includegraphics{#2}}}}

\AtBeginDocument{\def\includegraphics{\@ifnextchar[{\ts@includegraphics}{\ts@includegraphics[width=\checkGraphicsWidth,height=\checkGraphicsHeight,keepaspectratio]}}}

\DeclareMathAlphabet{\mathpzc}{OT1}{pzc}{m}{it}

\def\URL#1#2{\@ifundefined{href}{#2}{\href{#1}{#2}}}

\edef\fntEncoding{\f@encoding}

\makeatother

\newif\ifmultipleabstract\multipleabstractfalse%
\usepackage[hmargin=1.8cm,vmargin=2.2cm]{geometry}
\makeatletter
\def\author#1{\gdef\@author{\hskip-\dimexpr(\tabcolsep)\hskip1pt\parbox{\dimexpr\textwidth-1pt}{\centering #1}}}

\let\@articletype\@empty \def\articletype#1{\gdef\@articletype{{\fontsize{14}{16}\selectfont #1}}}

\usepackage{fancyhdr}

\fancypagestyle{headings}{\fancyhf{}\fancyhead[RE]{\MakeTextUppercase{\textbf{\RunningAuthor}}}\fancyhead[LE]{\textbf{\thepage}}\fancyhead[LO]{\MakeTextUppercase{\textbf{\RunningHead}}}\fancyhead[RO]{\textbf{\thepage}}}
\fancypagestyle{plain}{\fancyhf{}\fancyfoot[C]{\textbf{\thepage}}}

\AtBeginDocument{\usepackage{lastpage}}
\def\title#1{%
  \gdef\@title{%
    \ifx\@articletype\@empty\else\@articletype~\\\fi%
     #1}%
}

\usepackage{abstract}
\def\abstractname{\textbf{Abstract}}
\renewenvironment{onecolabstract}
{\vspace*{-.4pc}\trivlist\item[]\leftskip1pt\noindent\selectfont\hfill\abstractname\hfill\mbox{\null}\par\ignorespaces}{\endtrivlist}

\def\NormalBaseline{\def\baselinestretch{1.1}}

\usepackage{textcase}
\usepackage[explicit]{titlesec}
\titleformat{\section}[block]{\NormalBaseline\boldmath\bfseries}
{\thesection.}
{6pt}
{#1}
[]
\titleformat{\subsection}[hang]{\NormalBaseline\filright\itshape}
{\thesubsection.}
{6pt}
{#1}
[]
\titleformat{\subsubsection}[hang]{\NormalBaseline\filright\itshape}
{\hspace{16pt}\thesubsubsection}
{6pt}
{#1}
[]
\titleformat{\paragraph}[runin]{\NormalBaseline}
{\theparagraph}
{6pt}
{#1}
[]
\titleformat{\subparagraph}[runin]{\NormalBaseline}
{\thesubparagraph}
{6pt}
{#1}
[]

\titlespacing{\section}{0pt}{1.5\baselineskip}{.2\baselineskip}  
\titlespacing{\subsection}{0pt}{1.5\baselineskip}{.2\baselineskip}  
\titlespacing{\subsubsection}{0pt}{1.5\baselineskip}{.2\baselineskip}  
\titlespacing{\paragraph}{0pt}{.5\baselineskip}{10pt}  
\titlespacing{\subparagraph}{0pt}{.5\baselineskip}{10pt}

\usepackage{caption}
\DeclareCaptionLabelFormat{figlabel}{Figure #2}
\date{}
\makeatother

\usepackage{endfloat}
\usepackage{float}

\begin{document}

\title{MAT-DiSMech: A Discrete Differential Geometry-based Computational Tool for Simulation of Rods, Shells, and Soft Robots}
\def\RunningHead{MAT-DisMech}
\def\RunningAuthor{Lahoti \etal}
\author{Radha Lahoti, M. Khalid Jawed
\thanks{Radha Lahoti, M. Khalid Jawed are with Department of Mechanical and Aerospace Engineering, University of California Los Angeles, Los Angeles, CA 90024, USA 
        {\tt\small radhalahoti@g.ucla.edu, khalidjm@seas.ucla.edu}}%
}

\maketitle


{\begin{onecolabstract}
Accurate and efficient simulation tools are essential in robotics, enabling the visualization of system dynamics and the validation of control laws before committing resources to physical experimentation. Developing physically accurate simulation tools is particularly challenging in soft robotics, largely due to the prevalence of geometrically nonlinear deformation. A variety of robot simulators tackle this challenge by using simplified modeling techniques -- such as lumped mass models -- which lead to physical inaccuracies in real-world applications. On the other hand, high-fidelity simulation methods for soft structures, like finite element analysis, offer increased accuracy but lead to higher computational costs. In light of this, we present a Discrete Differential Geometry-based simulator that provides a balance between physical accuracy and computational efficiency. Building on an extensive body of research on rod and shell-based representations of soft robots, our tool provides a pathway to accurately model soft robots in a computationally tractable manner. Our open-source MATLAB-based framework is capable of simulating the deformations of rods, shells, and their combinations, primarily utilizing implicit integration techniques. The software design is modular for the user to customize the code -- for example, add new external forces and impose boundary conditions-- to suit their requirements. The implementations for prevalent forces encountered in robotics, including gravity, contact, kinetic and viscous damping, and hydrodynamic and aerodynamic drag, have been provided. We provide several illustrative examples that showcase the capabilities and validate the physical accuracy of the simulator. The open-source code is available at \href{https://github.com/StructuresComp/dismech-matlab.git}{\tt \small https://github.com/StructuresComp/dismech-matlab}. We anticipate that the proposed simulator can serve as an effective digital twin tool, enhancing the Sim2Real pathway in soft robotics research.

\def\keywordstitle{Keywords}
\end{onecolabstract}}

\begin{multicols}{2}
\section{Introduction}

Soft robots are robotic systems designed to accomplish desired objectives through the controlled deformation of their flexible structures \cite{softRobotDefinition_Wang}. These robots are largely inspired by biological creatures who appear to effectively leverage their bodily softness for efficient movement in complex terrains. Soft robots aim to replicate the adaptability, agility, and resilience found in nature. In addition to this, such robots also lead to enhanced safety in human-robot interaction owing to their inherent compliance. 
To fully harness the potential of soft robots, robust and accurate computational models are essential for understanding and predicting their complex behaviors. 

Modeling enables simulation, allowing researchers to create a digital twin for validating designs and concepts in a safe and controlled virtual environment before extensive physical experimentation \cite{why_simulation_PNAS}. Moreover, simulations offer the ability to explore physical parameters beyond the limitations of current hardware, such as miniaturizing designs -- a capability particularly valuable in biomedical applications \cite{miniaturization_robotic_surgery}.

However, accurately modeling the dynamics of soft robots -- which, in principle, possess infinite degrees of freedom -- presents significant challenges \cite{modeling_challenges} due to their highly nonlinear behavior. Unlike rigid body dynamics, where nonlinearity arises primarily from rotational kinematics while the body itself remains undeformed, soft body dynamics exhibit nonlinearities stemming from both rotations and large deformations of the continuum. These nonlinearities originate from two sources: geometry and material properties. Geometric nonlinearity arises when slender structures undergo significant bending, twisting, or stretching, leading to a nonlinear relationship between displacements and strains. In contrast, material nonlinearity refers to the nonlinear relationship between stress and strain inherent to the material’s constitutive behavior. In this paper, we focus on addressing geometric nonlinearity.

A spectrum of methods ranging from simplistic lumped-mass approaches to high-fidelity tools like finite element analysis (FEA) have been employed to tackle this challenge~\cite{soft_robot_modeling_review}. These techniques exhibit a visible trade-off between accuracy and efficiency, i.e., more physically accurate models like FEA tend to lack computational efficiency, and simpler models tend to compromise on physical accuracy~\cite{soft_robot_modeling_review_weicheng}.

A relatively newer approach in this area has been the use of Discrete Differential Geometry (DDG), which is the study of discrete counterparts to smooth geometric surfaces, focusing on the discrete versions of concepts like curvature and geodesics~\cite{ddg}. This mathematical discipline has found extensive applications in computer graphics research for visually realistic simulation of hair and clothes. Unlike FEA, in DDG-based methods, the deformation is modeled by nonlinear ordinary differential equations (ODEs) instead of partial differential equations (PDEs) in weak form. These methods use simpler discretized elements; for example, when modeling rods, the individual discrete segments do not undergo bending deformation, effectively reducing the degrees of freedom of each segment. This reduction in complexity, along with a focus on local interactions between discrete segments, makes it computationally more efficient than FEA, which requires solving large global stiffness matrices. This approach has been adapted in modeling the mechanics of slender structures, where it is shown to achieve remarkable physical accuracy~\cite{PhysRevLettFlagella, Jawed2014Coiling}. Recognizing its potential, Huang et al.~\cite{Huang2020a} first employed DDG-based 2D beam elements, successfully capturing the planar deformations of soft robot structures. Building on this foundation, extending DDG-based methods to model the full complexity of 3D soft body dynamics presents a promising avenue for advancing soft robot modeling.

A wide variety of soft robotic systems, including pneumatic network (PneuNet) actuators \cite{PneuNet, PneuNet_soro, pneunet_analytical_Ext}, mobile robots \cite{snake_soro, snake_soro2, multigait} and soft robotic grippers \cite{soft_gripper} can be modeled as a combination of one or more slender rod-like structures. The Discrete Elastic Rod (DER) algorithm proposed by Bergou et al. \cite{Bergou2008} is a method that has proven to be a promising tool for physically accurate and fast simulations of elastic rods. This technique works especially well for efficient soft contact handling even in complex scenarios such as knot tying~\cite{Jawed2015Knots, choi_imc_2021}, hence making it a great choice for rod-like soft robot simulations.

In addition to slender rods, thin plate and shell structures are highly prevalent in soft robotics, and a large variety of soft robots can be modeled as a combination of rods and shells\cite{Rus2015_soft_robot_review}. Modeling soft shells has garnered the interest of the computer graphics community, especially due to its application to realistic cloth simulations. The initial attempts to model deformation in thin plates used the notion of bending at hinges, which are edges shared by two mesh triangles~\cite{Baraff_1998}. This technique was modified by Grinspun et al. \cite{grinspun2003} to inculcate 3-dimensional shells, i.e. shells with non-zero natural curvature. While highly effective for generating visually realistic simulations in computer graphics, this method exhibits a significant dependence on the mesh's shape and orientation \cite{grinspun2006}, which hinders its ability to achieve the physical accuracy essential in robotics. Grinspun et al.~\cite{grinspun2006} and Chen et al.~\cite{libshell} presented advanced modeling techniques for elastic shells that remedy this mesh-dependent behavior, making them a promising tool for simulating soft structures in robotics.

Our work is motivated by the potential of DDG-based techniques to model flexible body dynamics, along with the growing demand for effective simulators in soft robotics, as evidenced by the recent development of various tools.
The widely used robot simulation environments such as Gazebo~\cite{koenig2004gazebo} and Bullet/PyBullet \cite{coumans2021_pybullet} employ rigid body dynamics to model robot structures and, hence, are not suitable for soft body simulation. SoMo~\cite{graule2020somo}, a wrapper around PyBullet, facilitates the simulation of soft links by approximating them as rigid links connected with springs. While computationally efficient, such methods fail to deliver adequate physical accuracy especially when dealing with complex soft structures.
Another class of prominent simulation tools include SOFA~\cite{coevoet_2017_sofa_robot}, utilized in the widely recognized Soft Robotics Toolkit developed by Holland et al.~\cite{toolkit}, and CHRONO \cite{CHRONO}, which use FEA to model soft body dynamics. While accurate in dynamic modeling, these simulators come at the cost of a higher computational expense. Yet another class of more recent soft robot simulators include Elastica \cite{Naughton2021_elastica_rl} and SoroSim \cite{mathew2022sorosim}, which use geometric modeling techniques like Cosserat rod theory to model soft rod dynamics. Although these simulators achieve a good balance of accuracy and efficiency, their applicability is limited to modeling rod-like flexible structures. Furthermore, Elastica employs an explicit time integration scheme, which requires very small timesteps to converge, especially when solving stiff systems of ODEs. This highlights the need for a simulation framework that combines accuracy, efficiency, and versatility to accommodate a broader range of soft robotic systems.

\begin{figure}[h!]
    \centering
    \includegraphics[width=1\columnwidth]{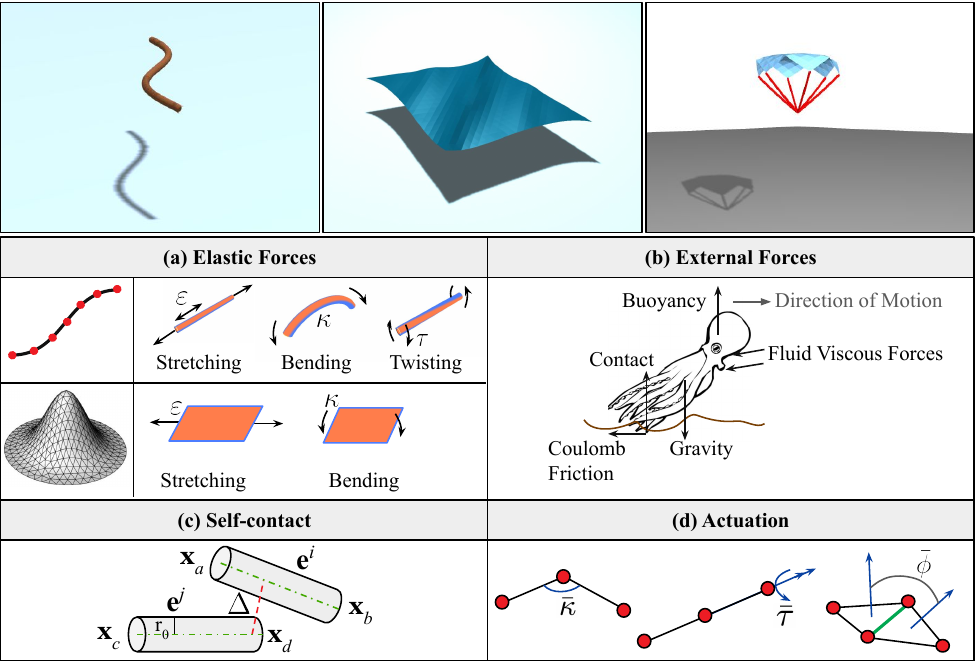}
    \caption{
    Examples of robots or soft structures: Snake, Manta ray, and Parachute, modeled using MAT-DiSMech. The key features within MAT-DiSMech.
    (a) \textbf{Elasticity:} The framework accounts for stretching, bending and twisting deformation modes for elastic rods and stretching, bending deformations for shells.
    (b) \textbf{External forces:} Some external forces implemented in MAT-DiSMech are shown. Additionally, the codebase allows users to implement their own external forces.
    (c) \textbf{Self contact:} Self contact is modeled using IMC \cite{choi_imc_2021, tong_imc_2022} which treats contact implicitly by defining a penalty energy as a function of $\Delta$, the minimum distance between two edges that can potentially come in contact.
    (d) \textbf{Actuation:} Actuation is primarily done by modifying the natural curvature or twist of the elastic elements. The parameters, $\bar{\kappa}, \bar{\tau}$, and $\bar{\phi}$, denote the natural curvature of the rod, the natural twist of the rod, and the natural hinge angle of the shell, respectively.} 
\label{fig:intro}
\end{figure}

In our previous work, we introduced a framework based on DDG for simulating multiple connected rod-like structures \cite{andrew}. The simulator coded in C++ leveraged the DER method to model rods and used an implicit integration scheme, showing an improvement in computational efficiency in comparison to the existing tools. Building on this foundation, we now present an enhanced generalizable simulation framework, which extends support to a wide range of soft robots composed of slender rods and thin shells. This new tool is developed fully in MATLAB and enables the simulation of a variety of soft robot configurations, such as one or more rods, shells, and their combinations, making it suitable for complex systems not easily addressed by existing methods. The MATLAB implementation, while somewhat slower than compiled languages, offers significant advantages in accessibility and usability by eliminating dependencies, reducing code complexity, and simplifying installation. Its open-source and dependency-free design makes it an excellent pedagogical tool for teaching and learning simulation methods. For researchers seeking greater computational efficiency, the code provides a clear and modular structure that can be adapted and reimplemented in a compiled language or integrated into existing architectures. Figure~\ref{fig:intro} shows the key features of MAT-DiSMech. The main contributions of our paper are as follows: 
\begin{enumerate}
    \item To our knowledge, MAT-DiSMech is the first DDG-based robot simulator capable of handling elastic rods, shells, and their combinations.
    \item We provide an open-source implementation of the discrete elastic shell bending energy that uses the discrete shape operator proposed by Grinspun et al. \cite{grinspun2006} that helps achieve higher mesh independence than hinge-based shell models.
    \item We provide an open-source MATLAB implementation of IMC\cite{tong_imc_2022}, an implicit contact modeling technique. This fully implicit method has proven highly effective in simulating elastic contact in contact-rich scenarios like knot tying and flagella bundling \cite{choi_imc_2021, tong_imc_2022}. 
\end{enumerate}

Our paper is structured as follows. Section~\ref{sec: dynamics} presents the kinematics and the equations of motion. Section~\ref{sec:elasticity} describes the formulation of the elastic energy of the entire structure. Section~\ref{sec: IMC_contact} covers the methods to model self-contact and Coulomb friction implicitly. Section~\ref{sec: external_forces} elaborates on the modeling of commonly encountered forces in robotics, in particular gravity, contact, coulomb and viscous friction, and aerodynamic drag. In Section~\ref{sec: overall_framework}, the overall framework of the simulator is described, and the procedure to go from a robot to its digital counterpart by providing appropriate inputs to the simulator is detailed. Finally, Section \ref{sec: results} showcases simulations of various soft robots and structures and results from physical validation of our solver.




\section{Dynamical Modeling}
\label{sec: dynamics}

A structure is discretized into $N$ nodes, which can represent a rod, a shell, or their connection. These nodes are used to define $E$ edges that form slender rods and $T$ triangles that constitute the shell structure. Each edge is a vector connecting two nodes, and each triangle is defined by three nodes connected together. Figure~\ref{fig:schematics}(a) shows a schematic of the discrete representation of a structure with $N=16$, $E=7$ and $T=10$. The user defines the geometry of the structure by providing the following information:
(i) the nodal position coordinates in 3D in the form of an $N\times 3$ array;
(ii) the array of size ($E\times 2$), of node indices $m$ and $n$ ($1 \le m, n \le N$) for all edges, $\mathbf e^i = \mathbf {x}_n - \mathbf {x}_m$, where $1 \le i \le E$ and 
(iii) the array of size ($T\times 3$) of node indices $l, m,$ and $n$ ($1 \le l, m, n \le N$), for each of the $T$ triangles.

\textit{Degrees of Freedom and Kinematics.} 
To model the dynamics of an elastic rod or a network of rods, we use the DER method~\cite{Bergou2008}, which builds upon the Kirchhoff rod theory~\cite{Kirchhoff1859}. In this framework, rods can undergo three primary deformation modes: stretching, bending, and twisting. While the original Kirchhoff rod model describes inextensible rods, the DER method extends this by introducing a stretching energy, allowing the rods to deform axially. If strict inextensibility is required, it is also possible to impose edge-length-preserving constraints, for instance, using the fast projection method~\cite{fast_projection}.

The degrees of freedom (DOF) vector for rods comprises the nodal positions, denoted by $\mathbf{x}$, and the twist angles of the edges, denoted by $\theta$, which capture their rotational state. Unlike rods, shells can undergo only two types of deformation: stretching or membrane deformation and bending. We implement two distinct discrete elastic shell theories to model shell bending, allowing the user to select their preferred method: (1) Hinge-based~\cite{grinspun2003} and (2) Mid-edge normal-based~\cite{grinspun2006}. For hinge-based bending, the DOF vector for the shell only consists of the nodal positions. Therefore, for a structure with $N$ nodes and $E$ edges, the DOF vector is of size $3N + E$ and is expressed as
\begin{align}
\label{eq: dof}
    \mathbf{q} = [\mathbf{x}_1,\mathbf{x}_2, ...,\mathbf{x}_{N},\theta^{1}, \theta^{2}, ..., \theta^{E}]^\top,
\end{align}
where superscript $^\top$ represents the transpose operator.
When using the mid-edge normal-based theory, it is necessary to track the edges on the shells -- similar to the edges on the rods -- by introducing an additional scalar degree of freedom, denoted by $\xi$, for each of the shell-edges. The definition of $\xi$ is explained in Section \ref{sec: midedge}. Given a shell with $T\times 3$ triangle array, the shell-edges can be obtained from the triangle array. Hence, if using the mid-edge normal-based formulation, assuming $Z$ to be the number of shell-edges obtained as above, the DOF vector is of size $3N+E+Z$ and is expressed as
\begin{align}
\label{eq: dof_midedge}
    \mathbf{q} = [\mathbf{x}_1,\mathbf{x}_2, ...,\mathbf{x}_{N},\theta^{1}, \theta^{2}, ..., \theta^{E}, \xi^{1}, \xi^{2}, ..., \xi^{Z}]^\top.
\end{align}

\textit{Equations of Motion.} 
The equations of motion (EOMs) of the system are
\begin{align}
\label{equ:eom}
    \mathbf{M}\Ddot{\mathbf{q}} - \mathbf{F}^{\text{elastic}} - \mathbf{F}^{\text{ext}} - \mathbf{F}^{\text{IMC}} = 0,
\end{align}
where $\mathbf{M}$ is a square diagonal lumped mass matrix of size $3N+E+Z$,
dot $\dot{(\,)}$ represents derivative with respect to time,
the vector $\mathbf F^\textrm{elastic} = - \frac{\partial E^{\textrm{elastic}}}{\partial \mathbf q}$ represents the elastic force (where $E^{\textrm{elastic}}$ is the elastic energy),
$\mathbf{F}^{\text{ext}}$ is the external force vector (such as gravity), and 
$\mathbf F^\textrm{IMC}$ represents the contact and friction forces following the implicit contact method (IMC)~\cite{tong_imc_2022, choi_imc_2021}.

The EOMs are solved numerically using the implicit Euler method. To step from time $t = t_k$ to $t = t_{k+1}$ with a time step size of $\Delta t = t_{k+1} - t_k$, we rewrite the EOMs as a system of first-order equations by introducing the velocity $\mathbf{u} = \dot{\mathbf{q}}$. The implicit Euler method then updates the state variables (positions and velocities) according to
\begin{align}
    \mathbf{f} \equiv \mathbf{M} \frac{1}{\Delta t} \left( \frac{\mathbf{q}_{k+1} - \mathbf{q}_k}{\Delta t} - \mathbf{u}_k \right) +  \frac{\partial E^\textrm{elastic}}{\partial \mathbf q_\textrm{k+1}} &\nonumber \\
    - \mathbf{F}^{\text{ext}}_{k+1} - \mathbf{F}^{\text{IMC}}_{k+1} = \mathbf 0 &, 
    \label{eq:position_update}\\
    \mathbf{u}_{k+1} = \frac{\mathbf{q}_{k+1} - \mathbf{q}_k}{\Delta t}&.
    \label{eq:velocity_update}
\end{align}
where the subscript $k$ and $k+1$ denotes evaluation of a quantity at time $t_k$ and $t_{k+1}$. Note that these equations are implicit in nature because the forces $\mathbf{F}^{\text{elastic}}_{k+1}$, $\mathbf{F}^{\text{ext}}_{k+1}$, and $\mathbf{F}^{\text{IMC}}_{k+1}$ depend on the unknown state at $t_{k+1}$. 

To solve Eq.~\ref{eq:position_update}, we employ the Newton-Raphson method and iteratively solve for $\mathbf{q}_{k+1}$. A line search method, similar to the one used in Tong el al.~\cite{tong_imc_2022}, is used to decide the step magnitude $\alpha$ in the direction of the gradient to help aid convergence, especially in the case of contact-rich scenarios. The running value of $\mathbf{q}_{k+1}$ is updated at every iteration as follows.
\begin{align}
\mathbf{q}_{k+1}^{(i+1)} = \mathbf{q}_{k+1}^{(i)} - \alpha\: \texttt{linearSolve}(\mathbf{J}, \mathbf{f}),
\end{align}
where $i$ is the iteration index, $\texttt{linearSolve}(\mathbf{J}, \mathbf{f})$ solves the linear system $\mathbf{J} \Delta \mathbf{q} = \mathbf{f}$ for $\Delta \mathbf{q}$, $\mathbf f$ is the residual of EOMs) evaluated at $\mathbf{q}_{k+1}^{(i)}$ from Eq.~\eqref{eq:position_update}, and $\mathbf{J}$ is the Jacobian matrix of the EOMs defined as
\begin{align}
\mathbf{J} = \mathbf{M} \frac{1}{\Delta t^2} + \frac{\partial^2 E^{\text{elastic}}}{\partial \mathbf{q}_{k+1}\partial \mathbf{q}_{k+1}} - \frac{\partial \mathbf{F}^{\text{ext}}}{\partial \mathbf{q}_{k+1}} - \frac{\partial \mathbf{F}^{\text{IMC}}}{\partial \mathbf{q}_{k+1}}.
\label{eq:Jacobian}
\end{align}
Here, $\mathbf{M} \frac{1}{\Delta t^2}$ represents the inertial contribution and is straightforward to compute; $\frac{\partial^2 E^{\text{elastic}}}{\partial \mathbf{q}_{k+1}\partial \mathbf{q}_{k+1}}$ is the Hessian of the elastic energy; $\frac{\partial \mathbf{F}^{\text{IMC}}}{\partial \mathbf{q}_{k+1}}$ accounts for the contribution of contact and friction forces; and $\frac{\partial \mathbf{F}^{\text{ext}}}{\partial \mathbf{q}_{k+1}}$ is the gradient of the external force. The last term can be ignored if their analytical expressions are not available, although doing so may reduce the convergence rate of the Newton-Raphson method. Once the positions $\mathbf{q}_{k+1}$ converge within a specified tolerance, the velocities $\mathbf{u}_{k+1}$ are updated using Eq.~\ref{eq:velocity_update}. The forces in Eq.~\ref{eq:position_update} -- arising from elasticity, contact/friction, and external effects -- and their gradients with respect to the DOFs in Eq.~\ref{eq:Jacobian} are described in the subsequent sections.

\section{Elastic Energy}
\label{sec:elasticity}
MAT-DiSMech decomposes the structure into a network of spring-analog elements, where each spring is associated with an elastic energy. These spring-like elements allow us to compute the total elastic energy of the structure by summing the contributions from all individual springs. A structure can have four types of springs, namely, stretching spring, bending-twisting spring, hinge spring, and triangle. The total elastic energy for $S$ stretching springs, $B$ bending-twisting springs, $H$ hinge springs and $T$ triangles  is
\begin{equation}\label{eq:E_elastic}
    \begin{split}
        E^{\textrm{elastic}} = \sum_{i=1}^S E^{\textrm{stretch}}_i + \sum_{i=1}^B (E^{\textrm{bend}}_i + E^{\textrm{twist}}_i) 
        + E^{\textrm{shell}},
    \end{split}
\end{equation}
where
\begin{equation*}
E^{\textrm{shell}} =
\begin{cases}
        \sum_{i=1}^H E^{\textrm{hinge}}_i & \textrm{\footnotesize{if hinge bending}} ,
        \\
        \sum_{i=1}^T E^{\textrm{mid-edge}}_i & \textrm{\footnotesize{if mid-edge bending}}. 
    \end{cases}
\end{equation*}
Stretching springs ($E^{\textrm{stretch}}_i$) can represent deformation in either rods or shells, while bending-twisting springs ($E^{\textrm{bend}}_i$ and $E^{\textrm{twist}}_i$) are exclusive to rods, capturing their bending and twisting behavior. Hinge springs and triangles are specific to shells, with the choice of hinge-based ($E^{\textrm{hinge}}_i$) or mid-edge normal-based ($E^{\textrm{mid-edge}}_i$) models determining the method used to model shell bending.

\subsection{Stretching Energy}\label{sec:2.1}
Stretching energy is stored in stretching springs associated with edges in rods and shells. For example, in Figure~\ref{fig:schematics}(b), the $i$-th stretching spring lies between nodes $\mathbf{x}_m$ and $\mathbf{x}_n$ on the edge $\mathbf{e}^i = \mathbf{x}_n - \mathbf{x}_m$. The axial strain in the $i$-th edge is given by
\begin{align}
\label{eq: stretch}
    \epsilon^{i} = \frac{\|\mathbf{e}^i\|}{\|\bar{\mathbf{e}}^i\|}-1,
\end{align}
where overbar $\bar{(\,)}$ represents a quantity in underformed or stress-free state and $\|\bar{\mathbf{e}}^i\|$ is the undeformed length of the $i$-th edge. The stretching energy along edge $\mathbf{e}^i$ is 
\begin{align}
\label{equ:stretching_energy}
    E^{\textrm{stretch}}_i = \frac{1}{2}k_s (\epsilon^{i})^2 \|\bar{\mathbf{e}}^{i}\|,
\end{align}
where $k_s$ is the stretching stiffness. For edges on rods, $k_s = EA$, where $E$ is the Young's modulus and $A$ is the area of cross-section. The default value for $A$ for rod-edges used in the simulator is $\pi r_0^2$ assuming a circular cross-section of radius $r_0$. The default value of $k_s$ for shell-edges is $\frac{\sqrt{3}}{4} E h \|\bar{\mathbf{e}}^{i}\|$, where $E$ is Young's modulus and $h$ is the thickness of the shell~\cite{gut}. If desired, users can modify the stiffness values of individual stretching springs in the simulator. Furthermore, the stiffness of each spring does not have to be the same.

\subsection{Bending and Twisting Energy}
\begin{figure}[h!]
        \centering
	\includegraphics[clip, trim=0cm 1cm 0cm 0cm,width=1\columnwidth]{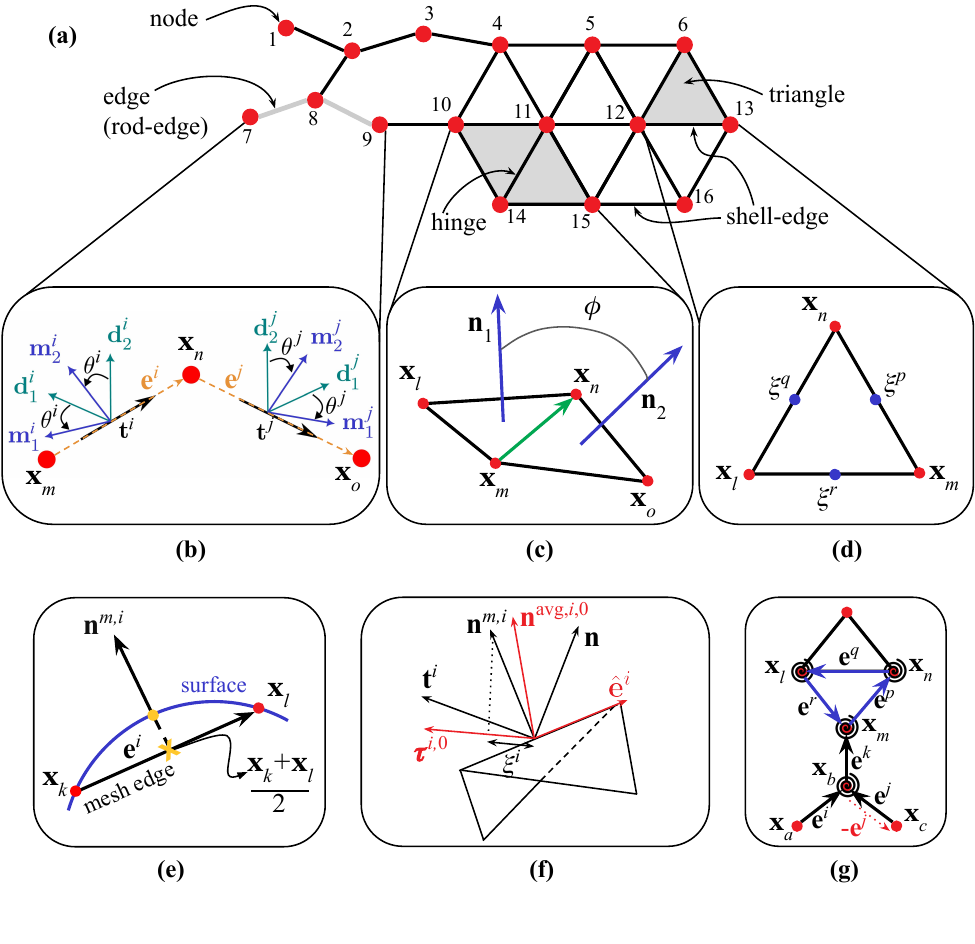}
	\caption{
    Schematics of physical modeling are shown here. (a) Discrete representation of a structure with both rod and shell elements. For this structure, the number of nodes, $N=16$, number of edges, $E=7$ and the number of triangles, $T=10$.  (b) Stencil of the bending-twisting spring for an elastic rod, comprising of nodes $\mathbf{x}_m$, $\mathbf{x}_n$, and $\mathbf{x}_o$ and edges $\mathbf{e}^i$ and $\mathbf{e}^j$ with attached reference frame ($\mathbf{a}_1, \mathbf{a}_2$) and material frame ($\mathbf{m}_1, \mathbf{m}_2$), used for the DER algorithm. (c) Schematic of a hinge spring for the hinge-based bending model for a discrete elastic shell. Hinge angle $\phi$ is used to calculate the bending energy. (d) The discretization stencil for the mid-edge normal-based bending energy model. For each shell-edge, $\xi^i$ is a scalar DOF that represents the rotation of the mid-edge normal about the edge. (e) Definition of the mid-edge normal. The blue curve labeled ``surface'' denotes the actual surface of the shell being modeled, ``mesh edge'' $\mathbf{e}$ denotes the edge of a triangle in the mesh that approximates the surface. The mid-edge normal $\mathbf{n}^m$ for $\mathbf{e}$ is normal to the surface, which, when extrapolated, intersects the triangle edge at its midpoint. (f) The schematic showcasing the edge attached reference frame $\{\mathbf{n}^{\textrm{avg},i},\mathbf{\tau}^i, \hat{\mathbf{e}}^i$\} and other vectors used in the mid-edge normal bending method \cite{grinspun2006}. (g) An elastic joint between two rods, and a rod and a shell. The node $\mathbf{x}_m$ shared between the rod-edge $\mathbf{e}^k$ and the shell triangle is denoted as a ``joint node'', and edges in blue color have the properties of both rod-edge and shell-edge. Bending-twisting springs are depicted by the spirals.} 
\label{fig:schematics}
\end{figure}

Bending and twisting energy for slender rods is stored in bending-twisting springs. A bending-twisting spring comprises three nodes - the center node at which the bending and twisting strains are measured and its two adjacent nodes, and two rod-edges - the edge entering the center node and the edge exiting the center node. In Figure~\ref{fig:schematics}(b),  three nodes ($\mathbf x_m, \mathbf x_n,$ and $\mathbf x_o$) and two edges ($\mathbf e^i$ and $\mathbf e^j$) form a bending-twisting spring. The edge $\mathbf e^i$ (and $\mathbf e^j$) is the vector from $\mathbf x_m$ to $\mathbf x_n$ (and from $\mathbf x_n$ to $\mathbf x_o$).
Each rod-edge $\mathbf{e}^i$ has two sets of orthonormal frames, a reference frame $\{\mathbf{d}^i_1,\mathbf{d}^i_2,\mathbf{t}^i\}$ and a material frame $\{\mathbf{m}^i_1,\mathbf{m}^i_2,\mathbf{t}^i\}$. Both of these frames share the tangent vector along the edge $\mathbf{t}^i=\mathbf{e}^i/\|\mathbf{e}^i\|$ as one of the directors. The reference frame is initialized at time $t=0$ and updated at each time step by parallel transporting it from the previous time step. The use of so-called time-parallel transport is a key feature of DER, leading to high computational efficiency~\cite{bergou2010discrete}. The material frame for the edge can be obtained from the reference frame by applying the twist angle for the edge, $\theta^i$, about their shared director $\mathbf{t}^i$ along the edge. 

Bending strain is measured at the middle node $\mathbf{x_n}$ through the curvature binormal vector
\begin{align}
\label{equ:curvature}
    (\boldsymbol \kappa b)_k = \frac{\text{2}\mathbf{e}^{i}\times \mathbf{e}^j}{\|\mathbf{e}^{i}\|\|\mathbf{e}^{j}\|+\mathbf{e}^{i}\cdot \mathbf{e}^{j}}.
\end{align}
The scalar curvatures along the first and second material directors using the curvature binormal are 
\begin{align}
\label{equ:curvature_material}
    \kappa^{(1)}_k = \frac{1}{2}(\mathbf{m}^{i}_2+\mathbf{m}^{j}_2)\cdot (\boldsymbol \kappa b)_k, \\
    \kappa^{(2)}_k = -\frac{1}{2}(\mathbf{m}^{i}_1+\mathbf{m}^{j}_1)\cdot (\boldsymbol \kappa b)_k.
\end{align}
The bending energy for the $k$-th bending-twisting spring is 
\begin{align}
\label{equ:bending_energy}
    E^{\textrm{bend}}_k = \frac{1}{2} \frac{EI}{\Delta l_k} [(\kappa^{(1)}_k - \bar{\kappa}^{(1)}_k)^2 + (\kappa^{(2)}_k - \bar{\kappa}^{(2)}_k)^2 ],
\end{align}
where $\Delta l_k = \frac{1}{2} (\|\bar{e}^{i}\| + \|\bar{e}^{j}\|)$ is the Voronoi length for $\mathbf x_k$, $\bar{\kappa}^{(1)}_k$ and $\bar{\kappa}^{(2)}_k$ are the natural curvatures in undeformed state, and $EI$ is the bending stiffness. The default value of $I$ is  $\pi r_0^4 /4$, assuming a circular cross-section of radius $r_0$.

The twist between the two edges $i$ and $j$ corresponding to the $k$-th bending-twisting spring is 
\begin{align}
\label{equ:twist}
    \tau_k = \theta^j - \theta^{i} + \Delta m_{k,\textrm{ref}},
\end{align}
where $\Delta m_{k,\textrm{ref}}$ is the reference twist, which is the twist of the reference frame as it moves from the $i$-th edge to the $j$-th edge. The twisting energy for the $k$-th bending-twisting spring is 
\begin{align}
\label{equ:twisting_energy}
    E^{\textrm{twist}}_k = \frac{1}{2} \frac{GJ}{\Delta l_k} (\tau_k - \bar{\tau}_k)^2,
\end{align}
where $\bar{\tau}_k$ is the twist along the centerline in the undeformed state, $\Delta l_k$ is the Voronoi length for the center node, $GJ$ is the twisting stiffness, $G$ is the Shear modulus of the rod's material given by $G = E/(2(1+\nu))$, $\nu$ is the Poisson's ratio of the rod's material and $J$ is the polar moment of inertia. The default value of $J$ is  $\pi r_0^2 /2$, assuming a circular cross-section. As in the case of stretching springs, if desired, users can modify the bending or twisting stiffness values of individual bending-twisting springs in the simulator.

An edge may be shared by multiple bending-twisting springs, especially while modeling a network of rods. The above formulation assumes a specific orientation for the edges ($\mathbf{x}_m \to \mathbf{x}_n$ for the first edge and $\mathbf{x}_n \to \mathbf{x}_o$ for the second edge). As shown in Figure~\ref{fig:schematics}(g), for bending-twisting spring $\{\mathbf{x}_a, \mathbf{x}_b, \mathbf{x}_c \text{ and } \mathbf{e}^i, \mathbf{e}^j\}$, since both rod-edges $\mathbf{e}^i$ and $\mathbf{e}^j$ are pointing towards the node $\mathbf{x}_b$, the negative of $\mathbf{e}^j$ is used for calculations so that it points away from and $\mathbf{e}^i$ points towards the node $\mathbf{x}_b$. The reference frame vector $\mathbf{d}^j_1$, material frame vector $\mathbf{m}^j_1$, and edge rotation angle $\theta^j$ are multiplied by $-1$ for computing the bending and twisting forces. Additionally, the resulting force vector and the Jacobian are reoriented to their original states by multiplying the affected terms by $-1$ if the edge and its related quantities were adjusted during the computation. Note that we could have equivalently flipped the edge $\mathbf{e}^i$ instead of $\mathbf{e}^j$ and proceeded as above, and the result would be mathematically equivalent. In the simulator, bending-twisting springs are initialized during the problem setup, and information about whether an edge needs to be negated is stored at the beginning of the simulation. This straightforward bookkeeping approach enables efficient handling of rod-like network structures.

\subsection{Shell Bending Energy}
Two different methods are available in MAT-DiSMech to formulate the bending energy for an elastic shell. The hinge-based formulation~\cite{grinspun2003}, while simple, leads to mesh-dependent behavior, meaning that the mesh orientation significantly influences the deflection. In contrast, the mid-edge normal-based method~\cite{grinspun2006} yields more mesh-independent behavior, making it more appropriate for applications that require higher accuracy.

\subsubsection{Hinge-based Bending Energy}
This method was initially proposed by Baraff et al.~\cite{Baraff_1998} to simulate plates, and was further modified by Grinspun et al. \cite{grinspun2003} to include shells with non-zero natural curvatures.
Referring to Figure~\ref{fig:schematics}(c), a hinge is defined as any shell-edge that is shared by two adjacent shell triangles about which the two faces can rotate relative to each other. The hinge spring comprises four nodes ($\mathbf x_l, \mathbf x_m, \mathbf x_n,$ and $\mathbf x_o$), and two of these nodes define the hinge, which in this case is the edge vector from $\mathbf x_m$ to $\mathbf x_n$. The hinge angle $\phi$ is defined as the angle between the normal vectors on these two shell triangles. 
The bending energy corresponding to the $i$-th hinge is
\begin{equation}
\label{eq:bending_energy_shell}
    E^{\textrm{hinge}}_i = \frac{1}{2} k_b \left(\phi_i - \bar{\phi}_i\right)^2,
\end{equation}
where $k_b = \frac{1}{\sqrt{3}} \frac{E h^3}{12}$ is the bending stiffness~\cite{gut}, $E$ is the Young's modulus and $h$ is the thickness of the shell. 

\subsubsection{Mid-edge Normal-based Bending Energy}
\label{sec: midedge}
In this method, the notion of the shape operator is used to describe curvature. The shape operator, denoted as $\Lambda_\mathbf{p}$, at a point $\mathbf{p}$ on a smooth surface, is a linear map that relates the tangent vector $\mathbf{t}$ at $\mathbf{p}$ to the rate of change of the surface normal vector $\mathbf{n}$ at $\mathbf{p}$ in the direction of $\mathbf{{t}}$ (see Grinspun et at.\cite{grinspun2006}). Mathematically, this relationship is expressed as
\begin{equation}
\label{eq: lambda_basic}
    \Lambda_\mathbf{p} \mathbf{t} = -\partial_\mathbf{t} \mathbf{n},
\end{equation}
where $\partial_\mathbf{t} \mathbf{n}$ denotes the directional derivative of $\mathbf{n}$ with respect to $\mathbf{t}$ that quantifies how the surface normal $\mathbf{n}$ changes in the direction of $\mathbf{t}$. The eigenvectors and eigenvalues of the shape operator at each point determine the directions in which the surface bends at that point.

To compute the discrete shape operator for a meshed surface, the concept of mid-edge normal is introduced. As shown in Figure~\ref{fig:schematics}(e), for a particular edge $\mathbf{e}^i$ on the mesh, the mid-edge normal $\mathbf{n}^{m,i}$ for that edge is defined as the normal to the smooth surface (approximated by the mesh), which, when extrapolated, intersects the edge at its mid-point. In Figure~\ref{fig:schematics}(f), a reference frame \{$\mathbf{n}^{\textrm{avg},i},\mathbf{\tau}^i, \hat{\mathbf{e}}^i$\} is attached to edge $\mathbf{e}^i$. Here, $\mathbf{n}^{\textrm{avg},i}$ is defined as the unit vector along $\frac{\mathbf{n}_1 + \mathbf{n}_2}{2},$ where $\mathbf{n}_1$ and $\mathbf{n}_2$ are the unit normal vectors of the two faces that share the edge $\mathbf{e}^i$, $\hat{\mathbf{e}}^i$ is the unit vector along $\mathbf{e}^i$ and $\boldsymbol{\tau}^i = \mathbf{n}^{\textrm{avg},i} \times \hat{\mathbf{e}}^i$. At the start of a time step, $\boldsymbol{\tau}^i $ is computed for each shell-edge, and this quantity is denoted as $\boldsymbol{\tau}^{i,0}$. For the $i$-th edge, a scalar $\xi^i$ is defined as the projection of the mid-edge normal on $\mathbf{\tau}^{i,0}$, i.e. $\xi^i = \mathbf{n}^{m,i} \cdot \boldsymbol{\tau}^{i,0}$ in Figure~\ref{fig:schematics}(f). This scalar $\xi^i$ is considered as a DOF associated with each shell-edge that represents the rotation of the mid-edge normal about the edge. 
Then, if a triangle has three nodes with indices $\{l,m,n\}$ and three edges with indices $\{p,q,r\}$ as shown in Figure~\ref{fig:schematics}(d), we have a DOF vector of size $(12\times 1)$,  $\mathbf{q}_{\textrm{mid-edge}} = \left[\mathbf{x}_l^\top, \mathbf{x}_m^\top, \mathbf{x}_n^\top, \xi^p, \xi^q, \xi^r\right]^\top$. We use the following per-triangle discrete shape operator presented by Grinspun et al.~\cite{grinspun2006} to compute the bending energy associated with the $i$-th triangle,
\begin{align}
\label{eq:lambda}
    \Lambda_i = \sum_{k} \frac{s^k\xi^k - (\mathbf{n}_i\cdot{\boldsymbol{{\tau}}^{k,0}})}{\bar{A}_i \|\mathbf{\bar{e}}^k\| (\hat{\mathbf{t}}^k \cdot \boldsymbol{\tau}^{k,0})} \: \mathbf{t}^k \otimes \mathbf{t}^k
\end{align}
where $k \in \{p,q,r\} $ denotes the $k$-th edge of the triangle, $\mathbf{n}_i$ denotes the unit normal to the triangle, $\bar{A}_i$ denotes the undeformed area of the triangle, $\|\mathbf{\bar{e}}^k\|$ denotes the undeformed length of the $k$-th edge, $\mathbf{t}^k$ is the tangent to the surface perpendicular to the edge (note that the $\hat{(\,)}$ operator indicates unit vector), $s^k$ is a variable that takes values $1$ or $-1$ depending on the ownership of the mid-edge normal, and the operator $\otimes$~denotes the outer product. Figure~\ref{fig:schematics}(e) shows all the above-defined vectors. Note that the ``shape operator'' defined in Eq.~\eqref{eq:lambda} is not operating on a point. Instead, it refers to the $(3\times3)$ matrix that represents the linear map from the shape operator definition in Eq.~\eqref{eq: lambda_basic}.
For the $i$-th triangle, the discrete shell bending energy, in terms of the discrete shape operator $\Lambda_i$, is
\begin{equation}
\label{eq: midedge_energy}
\begin{split}
    E^{\textrm{mid-edge}}_i = k_b \bar{A}_i[(1-\nu)\mathrm{Tr}((\Lambda_i-\bar{\Lambda}_i)^2) + \\
\nu(\mathrm{Tr}(\Lambda_i) - \mathrm{Tr}(\bar{\Lambda}_i) )^2],
\end{split}
\end{equation}
where $k_b = \frac{Eh^3}{24(1-\nu^2)}$, $E$ is the Young's modulus, $h$ is the thickness, $\nu$ is the Poisson's ratio of the shell, $\bar{A}_i$ and $\bar{\Lambda}_i$ denote the area and the shape operator of the $i$-th triangle in the undeformed or natural configuration, and $\mathrm{Tr}(\,)$ denotes the trace of a matrix. 
The complete analytical expressions for the gradient and Hessian of the energy in Eq.~\ref{eq: midedge_energy} are not available in previous publications, but are needed in Eqs.~\ref{eq:position_update} and~\ref{eq:Jacobian}; hence, we provide the complete expressions in Appendix~\ref{AppendixA:Midedge}. 

\subsection{Bending and Twisting at the Rod Shell Joint}
A rod-shell joint is formed when a node is shared between an  edge of a rod and one or more shell triangles, which we refer to as the "joint node." In Figure~\ref{fig:schematics}(f), $\mathbf x_m$ is such a node. To model its elastic energy, all stretching, bending, and twisting deformations involving the joint node must be considered. For this purpose, we treat the edges ($\mathbf e^p = \mathbf x_n - \mathbf x_m$, $\mathbf e^q = \mathbf x_l - \mathbf x_n,$ and $\mathbf e^r = \mathbf x_m - \mathbf x_l$) that form the triangles sharing the joint node as if they are also rod-edges, augmenting them with material and reference frames and associated twist angles. As illustrated in Figure~\ref{fig:schematics}(g), bending and twisting springs are then associated with every possible three-node, two-edge combination resulting from this assumption, i.e., $\{\mathbf x_b, \mathbf x_m, \mathbf x_n~\textrm{and}~\mathbf{e}^k, \mathbf{e}^p\}$, $\{\mathbf x_m, \mathbf x_n, \mathbf x_l~\textrm{and}~\mathbf{e}^p, \mathbf{e}^q\}$, $\{\mathbf x_n, \mathbf x_l,\mathbf x_m~\textrm{and}~\mathbf{e}^q, \mathbf{e}^r\}$, $\{\mathbf x_l, \mathbf x_m,\mathbf x_n~\textrm{and}~\mathbf{e}^r, \mathbf{e}^p\}$, and $\{\mathbf x_l,\mathbf x_m, \mathbf x_b~\textrm{and}~\mathbf{e}^r, \mathbf{e}^k\}$. Consequently, if the mid-edge normal-based bending method is used, the edges $\mathbf e^p, \mathbf{e}^q$ and $\mathbf{e}^r$ will possess both $\theta$ and $\xi$ degrees of freedom. The joint node, in turn, will experience elastic forces from bending and twisting of rods, bending of shells, and stretching of edges.



\section{Self-contact and Friction}
\label{sec: IMC_contact}
In addition to modeling inherent elasticity, we account for contact and friction forces resulting from the structure\rq{}s self-interactions for a realistic representation of its behavior during deformations and collisions. We do so by using a fully implicit penalty energy method, Implicit Contact Model (IMC)~\cite{choi_imc_2021, tong_imc_2022}, that integrates seamlessly into our framework. We divide the force into two parts: contact force to ensure non-penetration and friction forces due to Coulomb friction. 

The contact force is evaluated for each pair of non-adjacent edges that are in close proximity, denoted as a ``contact pair''. Figure~\ref{fig:intro}(c) shows a contact pair comprising of two edges $\{\mathbf{e}^i, \mathbf{e}^j\}$ or four nodes $\{\mathbf{x}_a$, $\mathbf{x}_b$, $\mathbf{x}_c$ , $\mathbf{x}_d\}$, since $\mathbf{e}^i = \mathbf{x}^b - \mathbf{x}^a$ and $\mathbf{e}^j = \mathbf{x}^d - \mathbf{x}^c$, and $\Delta (\geq 0)$ is the shortest distance between $\mathbf{e}^i$ and $\mathbf{e}^j$ measured using Lumelsky's algorithm~\cite{LUMELSKY198555}. Penetration occurs if $\Delta < 2r_0$. To compute the contact force for each pair, IMC defines a contact energy that penalizes penetration as follows,

\begin{equation}
\label{eq:contact_energy}
    E^{\textrm{con}} = 
    \begin{cases}
        (2r_0 - \Delta)^2 & \Delta \leq 2r_0-\delta, 
        \\
        0 &  \Delta \geq 2r_0 + \delta,
        \\
        \left(\frac{1}{K_1}\log\left(1 + e^{K_1 (2r_0 - \Delta)}\right) \right)^2  & \textrm{otherwise} 
    \end{cases}
\end{equation}
where $K_1 = 15/\delta$ is a stiffness parameter and $\delta$ is a user-defined contact distance tolerance. Similar to elastic energy, for a contact pair, the contact force on the $i$-th node of the contact pair is derived as the gradient of the contact energy with respect to the degrees of freedom of the node as,
\begin{equation}
    \mathbf{F}_i^\textrm{con} = -\dfrac{\partial E^{\textrm{con}}}{\partial \Delta} \dfrac{\partial \Delta}{\partial \mathbf q_i}.
\end{equation}



If friction in non-zero, Coulomb friction force on the $i$-th node of the contact pair arising due to the contact force  $\mathbf{F}_i^\textrm{con}$ is
\begin{align}
    \mathbf F_i^\textrm{fr} &= - \mu \gamma \hat{\mathbf u} \lVert \mathbf F^{\textrm{con}}_i \rVert, \nonumber \\
    \gamma &= \dfrac{2}{1 + e^{-K_2 \lVert \mathbf u \rVert}} - 1,
\label{eq: IMC_friction}
\end{align}
where $\mathbf u$ is the tangential relative velocity of the two edges in contact, which can be computed from the velocities of the nodes in the contact pair~\cite{tong_imc_2022}; $\mu$ is the friction coefficient; $K_2 = 15 / \nu$ is a stiffness parameter used to calculate $\gamma \in [0, 1]$ which is a smooth scaling factor that models the transition from sticking and sliding modes between the contacting bodies; here $\nu$ is a user-defined slipping tolerance, i.e., the velocity beyond which the bodies are considered to undergo slipping. The total force due to self-contact and friction is obtained by summing over the contribution from all contact pairs.
\begin{equation}
    \mathbf{F}^{\textrm{IMC}} = \sum_{\{a,b,c,d\}\in\textrm{contact~pair}}\left( \sum_{i\in\{a,b,c,d\}}  \mathbf F^{\textrm{con}}_i + \mathbf F^{\textrm{fr}}_i \right).
\end{equation}

The expressions for the Jacobian of contact and friction forces are omitted here but can be found in Ref.~\cite{tong_imc_2022}. The above contact model is designed for soft contact between slender rods. However, the same code can be used by approximating shell triangles using their edges. For greater accuracy, the minimum distance ($\Delta$) can be defined as the shortest distance between two triangles instead of two edges, which can then be used to formulate the contact energy. The extension to accurately model shell contact is left for future work.


\section{External forces}
\label{sec: external_forces}
In addition to the elastic forces and self-contact (and friction), the algorithm models the commonly encountered environmental forces in robotics such as gravity, buoyancy, contact, coulomb and viscous friction, and aerodynamic drag. Moreover, since external forces are usually problem-dependent, the framework allows users to add custom external forces to the system as explained in section \ref{sec: custom_ext_f}. The external force formulations are provided per node, and the total external force is calculated by summing over all the nodes in the structure. 

\subsection{Gravity and Buoyancy}
If the mass associated with the $i$-th node is $M_i$ and $\mathbf{g}$ is the acceleration vector due to gravity, the gravity force on that node is $M_i \mathbf{g}$. To model the effect of the buoyancy force for a body submerged in a fluid medium, we replace the value of $\mathbf{g}$ by $\frac{\rho -\rho_{\textrm{med}}}{\rho} \mathbf{g}$, where $\rho$ is the density of the structure and $\rho_{\textrm{med}}$ is the density of the fluid medium.



\subsection{Contact and Friction with Ground}
When a node $\mathbf{x}_i$ comes in contact with the ground surface, i.e., the distance between the node and ground ($\Delta_i$) is less than some tolerance value ($\delta$), it will experience a repulsive force along the normal $\hat{\mathbf{n}}$ to the ground at the point of contact. This normal force encountered and its Jacobian is calculated using the following equations, respectively.
\begin{align}
\label{eq:floor_contact}
    &\mathbf{F}^{\text{c}}_i = k_c \frac{-2 e^{-K_1 \Delta_i} \log(e^{-K_1 \Delta_i} + 1)}{K_1 (e^{-K_1 \Delta_i} + 1)} \hat{\mathbf{n}},\\
    &\mathbf{J}^{\text{c}}_{ii} = k_c \frac{2 e^{-K_1 \Delta_i} \log(e^{-K_1 \Delta_i} + 1) + 2e^{-2K_1 \Delta_i}}{(e^{-K_1 \Delta_i} + 1)^2} \hat{\mathbf{n}} \hat{\mathbf{n}}^\top,
    \label{eq:floor_contact_jacob}
\end{align}
where $k_c$ is the contact stiffness, which depends on the firmness of the ground one wants to model, and $K_1$ is a tunable parameter which we have set to $15/\delta$.

Coulomb friction force due to the ground is computed using the above contact force similar to friction force due to self contact (see Eq.~\eqref{eq: IMC_friction}), except the velocity ($\hat{\mathbf{u}}$) in this case, is the velocity of the node in contact with the ground along the tangent to the ground. By default, ground is assumed to be parallel to the $x-y$ plane; hence $\hat{\mathbf{n}}$ is $[0,0,1]^\top$. Note that the above penalty energy based method for ground contact force is especially useful to model contact when friction is non-zero. For frictionless contact, a simpler predictor-corrector approach can be taken similar to Ref.~\cite{MKJawed2014} that ensures no penetration into the ground.   

\subsection{Viscous Damping}

Viscous friction is modeled by applying a dissipative force to each node, proportional to the nodal velocity. The viscous force on $i$-th node at $n$-th time instant is given by
\begin{equation}
    \mathbf F^{\text{visc}}_i = - \eta \cdot \left( \frac{\mathbf x_i (t_{n}) - \mathbf x_i (t_{n-1})} {\Delta t} \right) \cdot \Delta l_i
\end{equation}
where $\eta$ (Pa $\cdot$ s) is the viscosity coefficient of the medium and $\Delta l_i$ is the Voronoi length of the node.
The Jacobian of this force is
\begin{equation}
    \mathbf J^{\text{visc}}_{ii} = - \frac{\eta}{\Delta t} \cdot \Delta l_i \cdot\mathbb I,
\end{equation}
where $\mathbb I$ is the square identity matrix of size $(3\times 3)$.


\subsection{\mbox{Hydrodynamic force using Resistive Force Theory}}
Resistive Force Theory (RFT)~\cite{rft_gray} is among the most widely used hydrodynamic models for describing interactions between low-Reynolds-number flows and slender structures. For the $i$-th node, the hydrodynamic force at $n$-th time instant is calculated by summing contributions from all edges connected to it, as follows,
\begin{equation}
    \mathbf F^{\text{RFT}}_i = \sum_{k\in E} -\frac{l^k}{2}\left[(C_t-C_n)\hat{\mathbf{e}}^k(\hat{\mathbf{e}}^k)^\top + C_n\mathbb{I}\right]\mathbf{u}_i,
\end{equation}
where $E$ is the set of indices of edges which are connected to the $i$-th node, $\mathbf{u}_i = \left( \frac{\mathbf x_i (t_{n}) - \mathbf x_i (t_{n-1})} {\Delta t} \right)$ is the velocity of the $i$-th node at $n$-th time instant, $\Delta t$ is the simulation time step,$\hat{}$ denotes unit vector, $C_t$ and $C_n$ are the tangential and normal resistive coefficients respectively,  and $l_k$ is the lengths of the $k$-th edge in undeformed state. Since each edge is shared by two nodes, $l^k$ is divided by $2$ to get the length corresponding to one of those. The negative sign denotes that the force acts in the direction opposite to the velocity component. 
The Jacobian of this force is

\begin{align}
\label{eq:JRFT}
\begin{split}
    \mathbf J^{\text{RFT}}_{ij} &= \frac{\partial \mathbf{F}^{\textrm{RFT}}_i}{\partial \mathbf{x}_j}
    \\
    &=  \sum_{k\in E} - \frac{l^k}{2}
    \left[(C_t-C_n) (\mathbf{u}_i^\top{\hat{\mathbf{e}}^k}) \frac{\partial \hat{\mathbf{e}}^k}{\partial x_j} 
    \right.
    \\
    & \quad + (C_t-C_n) \hat{\mathbf{e}}^k\left( (\hat{\mathbf{e}}^k)^\top \frac{\partial \mathbf{u}_i}{\partial x_j} + \mathbf{u}_i^\top \frac{\partial \hat{\mathbf{e}}^k}{\partial x_j} \right)
    \\
    & \quad + \left.C_n \frac{\partial \mathbf{u}_i}{\partial x_j}
    \right],
\end{split}
\end{align}
where $\mathbb I$ is the identity matrix of size $(3\times 3)$.
 Note that $\mathbf J^{\text{RFT}}_{ij}$ will be non-zero only for $j=i$ and the indices $j$ that correspond to the nodes adjacent to the $i$-th node. The expressions for 
$\frac{\partial \hat{\mathbf{e}}^k}{\partial \mathbf{x}_j}$ and $\frac{\partial \mathbf{u}_i}{\partial \mathbf{x}_j}$ are provided in the Appendix \ref{AppendixB:Visc}.

\subsection{Aerodynamic Drag}
Aerodynamic drag is a dissipative force proportional to the square of the nodal velocity and acts on the nodes contained in the shell structure only. The drag force on the $i$-th node is obtained by summing over all shell triangles that share the node and adding a force proportional to the square of the component of the node velocity ($\mathbf{u}_i$) along the normal of the triangle in the direction opposite to the velocity component. Thus, the drag force acting on $i$-th node at $n$-th time instant is given by
\begin{equation*}
    \mathbf F^{\text{drag}}_i = \sum_{k\in T} -\frac{ \rho_{\textrm{med}} C_D}{2} \frac{A_k}{3} \texttt{sgn}\left( \mathbf{u}_i\cdot \hat{\mathbf{n}}_k \right) \left( \mathbf{u}_i\cdot \hat{\mathbf{n}}_k \right) ^2 \hat{\mathbf{n}}_k
\end{equation*}
where $\rho_{\textrm{med}}$ denotes the density of the fluid medium in which the robot is operating, $\mathbf{n}_k$ denotes the normal vector of the $k$-th triangle, $T$ is the set of indices of triangles which share the $i$-th node, $\mathbf{u}_i = \left( \frac{\mathbf x_i (t_{n}) - \mathbf x_i (t_{n-1})} {\Delta t} \right)$ is the velocity of the $i$-th node at $n$-th time instant, $\Delta t$ is the simulation time step, $C_D$ is the dimensionless drag coefficient, $A_k$ is the area of the $k$-th triangle in its undeformed state, $\texttt{sgn}$ denotes the signum function, and the negative sign implies that the force acts in the direction opposite to the velocity component. Since each triangle is shared by three nodes, $A_k$ is divided by $3$ to get the area corresponding to one of those.
The Jacobian of this force is given by
\begin{align}
\label{eq: Jdrag}
\nonumber
    &\mathbf J^{\text{drag}}_{ij} = \frac{\partial \mathbf{F}^{\textrm{drag}}_i}{\partial \mathbf{x}_j} =  \sum_{k\in T} -\frac{ \rho_{\textrm{med}} C_D A_k}{6} \texttt{sgn}\left( \mathbf{u}_i\cdot \hat{\mathbf{n}}_k \right)\\
    &\left[2 \hat{\mathbf{n}}_k 
    (\mathbf{u}_i\cdot \hat{\mathbf{n}}_k )\left(\hat{\mathbf{n}}_k^\top \frac{\partial \mathbf{u}_i}{\partial \mathbf{x}_j} + \mathbf{u}_i^\top \frac{\partial \hat{\mathbf{n}}_k}{\partial \mathbf{x}_j}\right) + (\mathbf{u}_i\cdot \hat{\mathbf{n}}_k )^2\frac{\partial \hat{\mathbf{n}}_k}{\partial \mathbf{x}_j} \right]
\end{align}

Note that $\mathbf J^{\text{drag}}_{ij}$ will be non-zero only for $j=i$ and the indices $j$ that correspond to the nodes that share a triangle with the $i$-th node. The expressions for 
$\frac{\partial \hat{\mathbf{n}}_k}{\partial \mathbf{x}_j}$ and $\frac{\partial \mathbf{u}_i}{\partial \mathbf{x}_j}$ are provided in the Appendix \ref{AppendixC:Aero}.

\subsection{Custom External Force}
\label{sec: custom_ext_f}
If a user desires to apply an external force of their choice other than the ones implemented in our simulator, they can do so by creating a custom function to calculate the external force and its Jacobian similar to the functions for the provided external forces. The inputs to the function can be the DOF vector, the velocity vector, and any other system parameters as necessary, and the output would be the force, a vector of the same length as the DOF vector and the Jacobian, a square matrix of the same size as the DOF vector.  Note that even though our method uses the force and Jacobian of the force to propagate the dynamics forward, the Jacobian is used to provide a direction of search for the solution to the equation of motion at every time step and is not a requirement. Hence, if the external force to be applied doesn't have an analytical gradient, one can still apply just the force and simulate the dynamics. Having a Jacobian aids in the convergence of the Newton-Raphson iterations, and one can use automatic differentiation or numeric differentiation tools as well to find the Jacobian if needed.

\section{Overall Framework}
\label{sec: overall_framework}
\begin{figure}[h!]
        \centering
	\includegraphics[width=1\columnwidth]{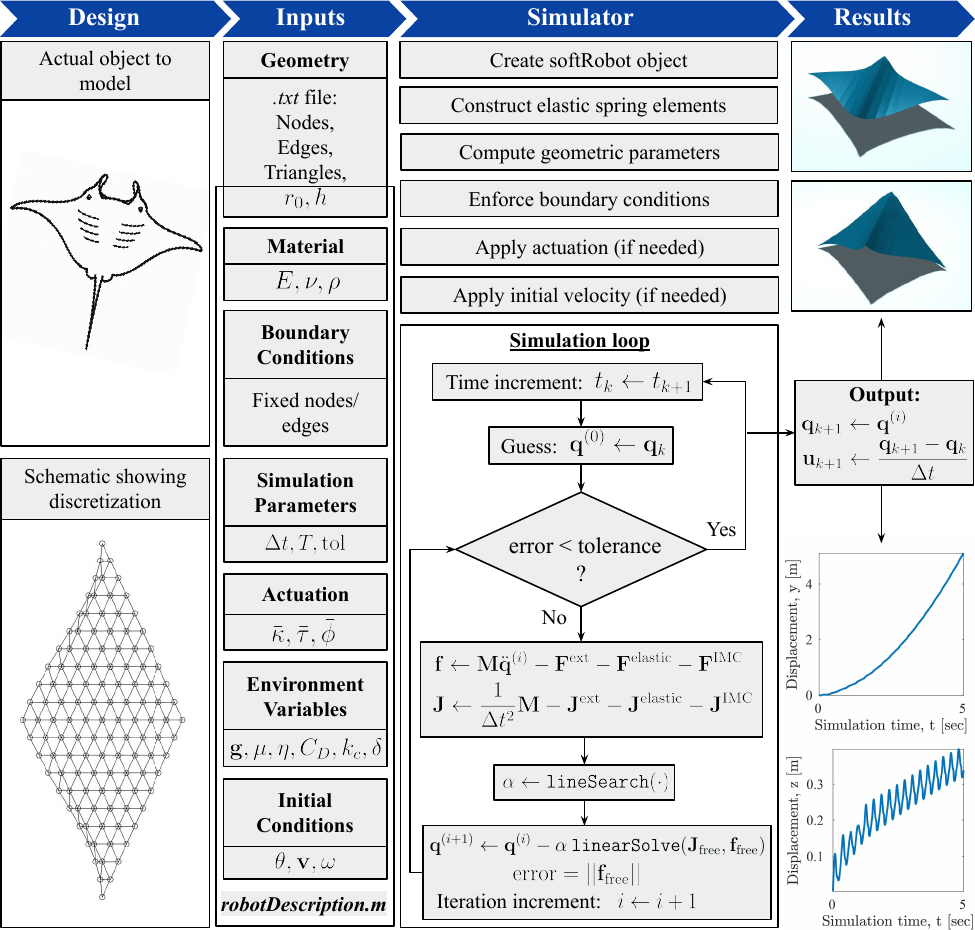}
	\caption{Overall framework of MAT-DiSMech. The first (leftmost) column shows how a physical structure can be modeled as a discrete shell structure. The second column details the inputs to be given to the simulator. The third column depicts the step-by-step simulation procedure. The fourth column represents the results in the form of simulation videos or plots of variation of useful physical quantities, such as displacements with time.} 
\label{fig:flowchart}
\end{figure}

The overall framework of our simulator is shown in Figure~\ref{fig:flowchart}.
\subsection{Input}
The input to the simulator consists of 

\noindent\textbf{Geometry: } A text (.txt) file specifying the geometric configuration of the soft structure. The file should contain 
    \begin{enumerate}
        \item the nodal position coordinates $(x, y, z)$,
        \item the edge array i.e., the node indices corresponding to each rod-edge, in case there are rod-like structures in the robot body. For example, in Figure~\ref{fig:schematics}(a), the edge array would be of size $(7\times 2)$ as follows,
        \[
\begin{bmatrix}
1 & 2 \\
2 & 3 \\
3 & 4 \\
7 & 8 \\
8 & 9 \\
9 & 10 \\
2 & 8
\end{bmatrix}
\]
        
        \item the triangle array for the shell, i.e., the node indices corresponding to each triangle, in case there is a shell structure in the robot body. This can be generated using MATLAB or any other external tool. For example, in Figure~\ref{fig:schematics}(a), the triangle array would be of size $(10\times 3)$ as follows,
        \[
\begin{bmatrix}
10 & 4 & 11 \\
4 & 5 & 11 \\
11 & 5 & 12 \\
12 & 5 & 6 \\
12 & 6 & 13 \\
10 & 11 & 14 \\
11 & 15 & 14 \\
11 & 12 & 15 \\
15 & 12 & 16 \\
12 & 13 & 16
\end{bmatrix}
\].
    \end{enumerate}  
    Other than the above, the dimensions of the cross-section of the rod and the thickness of the shell are to be provided. Some example input text files have been provided in the github repository for MAT-DiSMech.
    \vspace{0.25cm}
    
\noindent\textbf{Material Properties:}
    The material densities $\rho_{\text{rod}}, \rho_{\text{shell}}$, Young's modulii $Y_{\text{rod}}, Y_{\text{shell}}$, and poisson's ratios $\nu_{\text{rod}}, \nu_{\text{shell}}$ are inputs to the simulator.
    
    \vspace{0.25cm}
    
\noindent\textbf{Boundary Conditions:}
    The user will specify the fixed nodes and/or edges. Additionally one can also fix specific degrees of freedom for nodes, for instance, at a roller support, the motion is constrained in the $z$-direction only, hence to model this, the $x,y$ DOFs of the node can be left free, and $z$ fixed. 
    
    \vspace{0.25cm}
    
\noindent\textbf{Simulation Parameters:}
    The user specifies the time step $(\Delta t)$, the total time of simulation, and tolerance on the acceptable error while solving the Newton Raphson iterations. Additionally, there are flags to decide if the simulation is static or dynamic, if the simulation is 2D or 3D, if line search is enabled to aid in convergence, and if the shell simulation uses the hinge-based bending or the mid-edge based bending model. Logging and visualization methods are also provided, and the user can decide the frequency at which the data is saved, and the visualization is plotted in MATLAB.
    
    \vspace{0.25cm}
    
\noindent\textbf{Actuation:}
    Actuation is achieved by modifying the natural curvature/twist. To do so, the user can provide an Excel (.xls) or a text (.txt) file with time-varying values of natural curvature/twist and use the helper functions provided in the framework to apply them at every time step. We have provided a few examples of simulating actuated rod and shell structures in the GitHub repository. 

    \vspace{0.25cm}
    
\noindent\textbf{Environment and Contact Variables:}
    The environment variables are used to apply external forces to the system. The user provides the list of external forces that are to be used in the simulation. The different variables required for the forces implemented within the simulator are
    \begin{enumerate}
        \item Gravity: acceleration due to gravity ($\mathbf{g}$),
        \item Buoyancy: density of the medium $\rho_{\textrm{med}}$
        \item Contact: contact stiffness ($k_c$) and contact distance tolerance ($\delta$), for self-contact and contact with the ground,
        \item Friction: friction coefficient ($\mu$) for the structure's own material for self-contact as well as friction coefficient with the floor,
        \item Viscous friction: viscosity coefficient $\eta$,
        \item Aerodynamic drag: drag coefficient $C_D$ and density of the medium $\rho_{\textrm{med}}$
    \end{enumerate}

    \vspace{0.25cm}
    
\noindent\textbf{Initial Values:} Initial values can be specified for twist angle for rod-edges ($\theta$), velocity for nodes ($\mathbf{v}$), angular velocity for rod-edges or shell hinge angles ($\omega$).

\subsection{Time Stepping}
We employ the backward Euler method, an implicit time integration scheme that enables the use of larger time steps compared to explicit methods. However, backward Euler introduces significant artificial numerical damping, which can overly suppress vibrations in the structure. To address this, we also offer an implicit midpoint method, which helps reduce numerical damping and is preferable when accurate vibration behavior is important for a specific application.

Implicit integration provides substantial computational advantages over explicit schemes, particularly for systems with higher Young’s moduli. For context, Choi et al.~\cite{andrew} present a comparison of computational times between simulators employing explicit and implicit time-stepping methods. Nevertheless, for very simple systems with low Young’s moduli, explicit methods such as forward Euler can be faster, as they avoid the computational overhead of computing and inverting Jacobians. Therefore, we also include an implementation of the forward Euler method for users who prefer an explicit scheme.

Our framework allows users to easily replace the integration scheme to suit their requirements. It is worth noting that in contact-rich simulations involving soft robotic structures, the implicit solver may sometimes struggle to converge. To improve convergence in such scenarios, we provide an optional line search strategy that determines an appropriate step size, ensuring the error decreases with every iteration.

The complete pseudocode for MAT-DiSMech is given in Algorithm \ref{der_algo}.

\noindent
\begin{algorithm}[H]
\caption{MAT-DiSMech}
\begin{algorithmic}[1]
\label{der_algo}
    \STATE \textbf{Input:} $\mathbf{q}_k, {\mathbf{u}}_k$, $\Delta t$, {tolerance}
    
    \STATE \textbf{Output:} $\mathbf{q}_{k+1}, {\mathbf{u}}_{k+1}$
    
    \STATE \textbf{Require:} boundary conditions $\rightarrow$ free
    
    \STATE \textbf{Function} \FuncSty{timeStepper}:
    
    \STATE \hspace{\algorithmicindent} Guess: $\mathbf{q}^{(0)} \gets \mathbf{q}_k$
    
    \STATE \hspace{\algorithmicindent} $ i \gets 0$, $\epsilon \gets \infty$
    
    \STATE \hspace{\algorithmicindent} \textbf{while} $\epsilon > \textrm{tolerance}$ \textbf{do}

    \STATE \hspace{\algorithmicindent} \hspace{\algorithmicindent}
    \tcp{\textrm{prepare for iteration}}
    
    \STATE \hspace{\algorithmicindent} \hspace{\algorithmicindent} \tcp{\textrm{calculate} $\mathbf{F}^{\mathrm{elastic}}$ \textrm{and} $\mathbf{J}^{\mathrm{elastic}}$}

    \STATE \hspace{\algorithmicindent} \hspace{\algorithmicindent} \tcp{\textrm{calculate} $\mathbf{F}^{\mathrm{ext}}$ \textrm{and} $\mathbf{J}^{\mathrm{ext}}$}

    \STATE \hspace{\algorithmicindent} \hspace{\algorithmicindent} \tcp{\textrm{calculate} $\mathbf{F}^{\mathrm{IMC}}$ \textrm{and} $\mathbf{J}^{\mathrm{IMC}}$}

    \STATE \hspace{\algorithmicindent} \hspace{\algorithmicindent} $\mathbf{f}\gets \mathbf{M}\Ddot{\mathbf{q}}^{(i)} -\mathbf{F}^{\mathrm{ext}} - \mathbf{F}^{\mathrm{elastic}}-\mathbf{F}^{\mathrm{IMC}}$

    \STATE \hspace{\algorithmicindent}
    \hspace{\algorithmicindent} 
    $\mathbf{J}\gets \frac{1}{\Delta t^2} \mathbf{M} -\mathbf{J}^{\mathrm{ext}} - \mathbf{J}^{\mathrm{elastic}} - \mathbf{J}^{\mathrm{IMC}}$

    \STATE \hspace{\algorithmicindent}
    \hspace{\algorithmicindent} 
    $\mathbf{f}_{\mathrm{free}} \gets \mathbf{f}(\mathrm{free})$ 

    \STATE \hspace{\algorithmicindent}
    \hspace{\algorithmicindent} 
    $\mathbf{J}_{\mathrm{free}} \gets \mathbf{J}(\mathrm{free})$    

    \STATE \hspace{\algorithmicindent}
    \hspace{\algorithmicindent} 
    {$\Delta \mathbf{q}_{\mathrm{free}} \gets \FuncSty{linearSolve}(\mathbf{J}_{\mathrm{free}},\mathbf{f}_{\mathrm{free}})$}

    \STATE \hspace{\algorithmicindent} \hspace{\algorithmicindent}
    $\alpha \gets \FuncSty{lineSearch}(\cdot)$
    
    \STATE \hspace{\algorithmicindent}
    \hspace{\algorithmicindent} 
    $\mathbf{q}^{(i+1)}(\mathrm{free}) \gets \mathbf{q}^{(i)}(\mathrm{free}) - \alpha \Delta \mathbf{q}_{\mathrm{free}}$

    \STATE \hspace{\algorithmicindent}
    \hspace{\algorithmicindent} 
    $\epsilon \gets \lVert \mathbf{f}_{\mathrm{free}} \rVert$

    \STATE \hspace{\algorithmicindent}
    \hspace{\algorithmicindent} 
    $i \gets i+1$

    \STATE \hspace{\algorithmicindent} \textbf{end}

    \STATE \hspace{\algorithmicindent} $\mathbf{q}_{k+1} \gets \mathbf{q}^{(i)}$

    \STATE \hspace{\algorithmicindent} $\mathbf{u}_{k+1} \leftarrow (\mathbf{q}_{k+1} - \mathbf{q}_k) /\Delta t$
    \RETURN $\mathbf{q}_{k+1}, \mathbf{u}_{k+1}$
\end{algorithmic}
\end{algorithm}

\section{Results and Discussion}
\label{sec: results}
In this section, we present simulations of various rod-like and shell-like soft structures. We also compare the deflections obtained using our simulator with the analytical values for a cantilever beam. 
\vspace{-3mm}
\subsection{Simulation Experiments}
\begin{figure}[h!]
        \centering
\includegraphics[clip, trim=0cm 8.2cm 0cm 0cm, width=1\columnwidth]{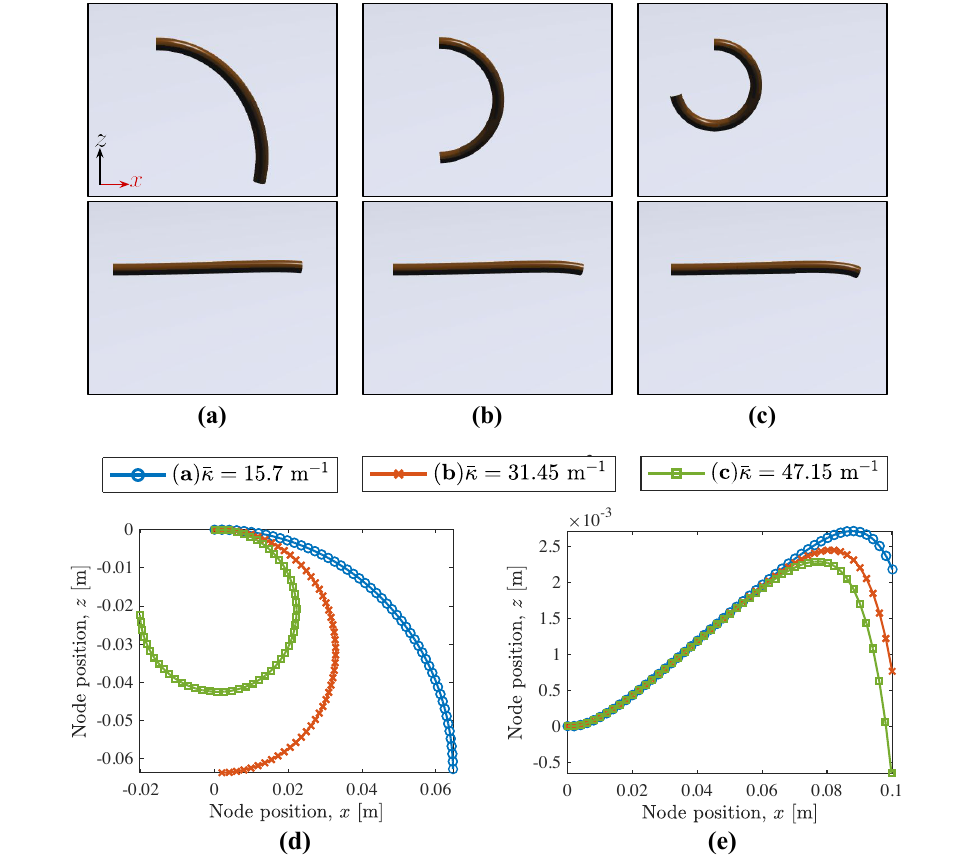}
 \caption{PneuNet actuator modeled as a discrete elastic rod. (a) Snapshots of the PneuNet actuator with natural curvature $15.7$ m$^{-1}$ with only gravity (top) and with gravity and point force on the free end (bottom). (b) Snapshots of the PneuNet actuator with natural curvature $31.45$ m$^{-1}$ with only gravity (top) and with gravity and point force on the free end (bottom). (c) Snapshots of the PneuNet actuator with natural curvature $47.15$ m$^{-1}$ with only gravity (top) and with gravity and point force on the free end (bottom). Note that the cross-sectional radius of the rod is $1$ mm, it has been enlarged in the figure for better visualization. (d) The plot of the natural shapes of the rods in (a), (b), and (c) under gravity. (e) Plot of the deformed shape of the rods under gravity and point force at the free end.}
\label{fig:rod_robot}
\end{figure}

\textit{PneuNet.}
We model a Pneumatic Network (PneuNet) actuator as a discrete elastic rod. The change in curvature upon application of pressure is modeled by modifying the natural curvature $\bar{\kappa}$ of the rod. Figure \ref{fig:rod_robot} shows three PneuNet actuators with natural curvatures of $15.70$ m$^{-1}$, $31.45$ m$^{-1}$, and $47.15$ m$^{-1}$ fixed at one end. Note that discrete curvature is dimensionless. The standard definition of curvature of the dimension 1/length relates to the discrete curvature for the $k$-th node as $\kappa_k/\Delta l_k$. Upon applying the same force at the end effector, these actuators take different shapes owing to their different natural curvatures. The deformed shapes of the actuators under a point force of $[175, 0, 7]^\top$ N at the free end and gravity with $\mathbf{g} = [0,0,-9.8]^\top$ m/s$^2$ are shown in Figure~\ref{fig:rod_robot}. We use $\rho = 1200 $ kg/m$^3, E = 20 $ GPa, and $\nu = 0.5$ as the material parameters for the system. Each rod is $0.1$ m long and has a cross-sectional radius of $1$ mm. This example is inspired by the work of Payrebrune et al. \cite{PneuNet}.

\begin{figure}[h!]
        \centering
	\includegraphics[clip, trim=0cm 8.2cm 0cm 0cm, width=1\columnwidth]{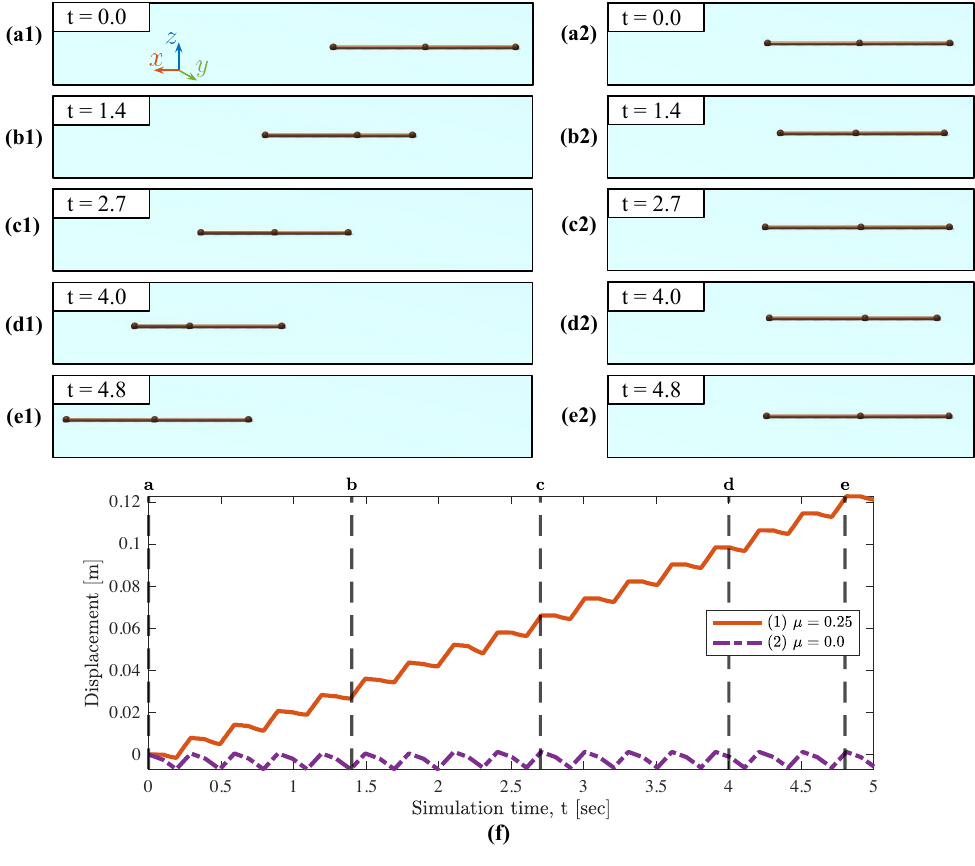}
 \caption{An earthworm modeled as an elastic rod in MAT-DiSMech. Snapshots of the earthworm's motion at different time stamps (a1-e1) with friction ($\mu = 0.25$), and (a2-e2) without friction. Here, time is given in seconds. (f) Displacement of the $x$-coordinate of the front end of the earthworm with time. Notice how, when there is no friction, the earthworm doesn’t move forward, and the contraction and expansion of the edges are also less pronounced compared to when there is friction. The time stamps corresponding to the snapshots (a-e) are marked in the plot using vertical dashed lines.}
\label{fig:earthworm}
\end{figure}

\begin{figure}[h!]
        \centering
	\includegraphics[clip, trim=0cm 10.7cm 0cm 0cm, width=1\columnwidth]{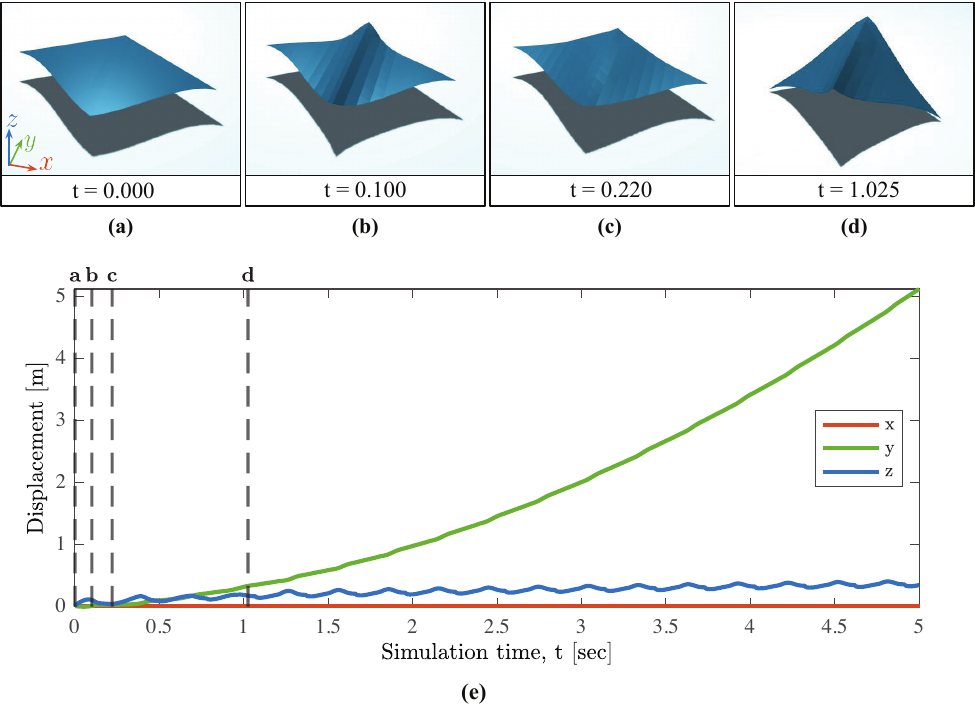}
 \caption{Manta ray modeled as an elastic shell in MAT-DiSMech. (a-d) Snapshots of the Manta ray's motion at different time stamps. Here the time is given in seconds. (e) Displacement of the mid-point of the leading edge with time. The time stamps corresponding to the snapshots (a-d) are marked in the plot using vertical dashed lines.}
\label{fig:mantaRay}
\end{figure}

\textit{Earthworm.}
As our second example, we simulate rectilinear locomotion in earthworms driven by axial peristaltic waves. We represent the worm as a single elastic rod composed of three nodes (forming two edges). Actuation is achieved by varying the natural length $(||\bar{\mathbf{e}}||)$ of the edges through alternating cycles of contraction and expansion. The contractions and expansions of the two edges are out of phase -- while one edge contracts, the other remains unchanged, and vice versa. This alternating pattern generates longitudinal asymmetry, allowing the worm to advance forward by exploiting frictional forces from the ground. Indeed, if there is no friction between the rod and the ground, the earthworm doesn't move forward and keeps oscillating in the longitudinal direction at its original position. This example is inspired by the work of  Rafsanjani et al.~\cite{snakeskin}. We use $E = 200$ MPa, $\nu = 0.5$ and $\rho = 1200$ Kg/m$^3$ and a cross-sectional radius of $0.001$ m for the rod. The length of the rod is $0.1$ m and each edge contracts/expands by $0.001$ m. The external forces acting on the earthworm are gravity, with $\mathbf{g} = [0,0, -9.8]^\top $ m/s$^2$ and frictional contact with the ground. The contact stiffness and coefficient of friction of the ground are 1000 and 0.25, respectively. Figure \ref{fig:earthworm} shows the intermediate poses of the earthworm during one actuation cycle and the trajectory of the front end when friction is present and when it is absent.
\textit{Manta ray.} 
For our third example, we simulate the flapping and undulating locomotion in a manta ray. We model the manta ray body as a discrete elastic shell by discretizing the system into an equilateral triangle mesh. The wingspan and the length of the ray are both $1$ m. The flapping and undulation are achieved through actuation at the center line. We modify the natural curvature angle ($\bar{\phi}$) of the hinges in a sinusoidal manner with a frequency of $3.4 $ Hz and an amplitude of $2\pi/3$ to achieve flapping. A phase lag of $\pi$ is introduced between the sine signals at the leading and trailing edge hinges, by varying the phase for the hinges on the center line linearly. We use $E = 6 $ GPa, $\nu = 0.3, \rho = 1057 $ kg/m$^3$ for the manta ray body. The external forces, in this case, are gravity, buoyancy, and aerodynamic drag, and we use $\mathbf{g} = [0,0,-9.8]^\top$ m/s$^2, \rho_{\textrm{med}} = 1000 $ kg/m$^3$ and $C_D = 0.5$. Figure~\ref{fig:mantaRay} shows the intermediate poses from the simulation of a manta ray in motion and the trajectory of the midpoint of its leading edge. It can be seen that the y-coordinate of the leading edge increases with time, indicating motion in the forward direction, the z-coordinate varies periodically, almost maintaining the altitude, and the x-coordinate doesn't change since the structure and the loads experienced are symmetric in the x-direction about the centerline. 

\begin{figure}[h!]
        \centering
	\includegraphics[clip, trim=0cm 7.1cm 0cm 0cm, width=1\columnwidth]{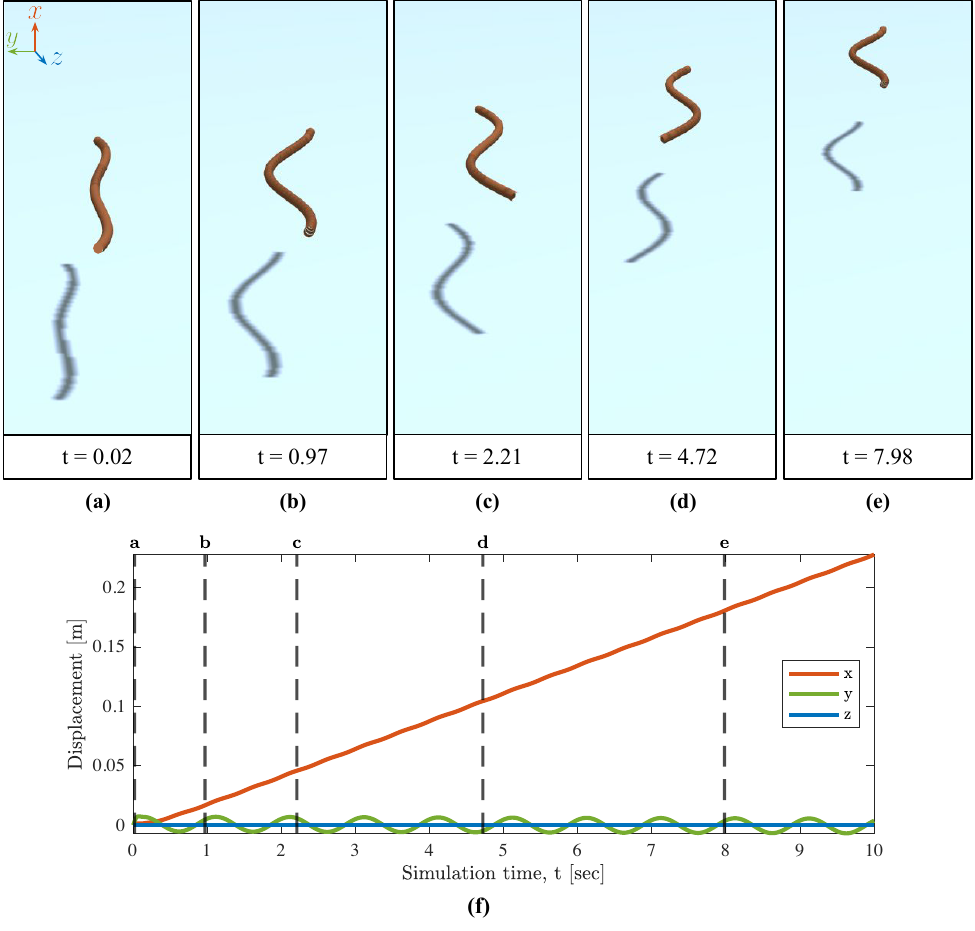}
 \caption{Snake modeled as an elastic rod in MAT-DiSMech. (a-e) Snapshots of the Snake's motion at different time stamps. Here the time is given in seconds. (f) Displacement of the leading end of the snake with time. The time stamps corresponding to the snapshots (a-e) are marked in the plot using vertical dashed lines.}
\label{fig:snake}
\end{figure}

\textit{Snake.}  
As our fourth example, we simulate the serpentine locomotion of a snake swimming in a fluid environment, subject to hydrodynamic forces. The snake is modeled as an elastic rod with a length of $0.1$~m, a cross-sectional radius of $1$~mm, Young’s modulus $E = 2$~MPa, Poisson’s ratio $\nu = 0.5$, and density $\rho = 1200$~kg/m$^3$. The surrounding fluid exerts hydrodynamic forces computed using resistive force theory (Gray et al. \cite{rft_gray}), with tangential and normal drag coefficients set to $0.01$ and $0.1$, respectively. This anisotropy in drag coefficients enables forward propulsion when serpentine undulations are applied. Serpentine motion is generated by modulating the natural curvature, $\bar{\boldsymbol{\kappa}}$, of the rod’s bending-twisting springs to produce a sinusoidal wave along the snake’s body, with an amplitude of $0.05$~m and a frequency of $1$~Hz. Figure~\ref{fig:snake} presents snapshots of the snake in motion and the trajectory of its leading end. The plots show that the $x$-coordinate steadily increases over time as the snake propels forward, the $y$-coordinate oscillates sinusoidally reflecting the serpentine path, while the $z$-coordinate remains constant at zero.

\begin{figure}[h!]
        \centering
	\includegraphics[clip, trim=0cm 5.6cm 0cm 0cm, width=\columnwidth]{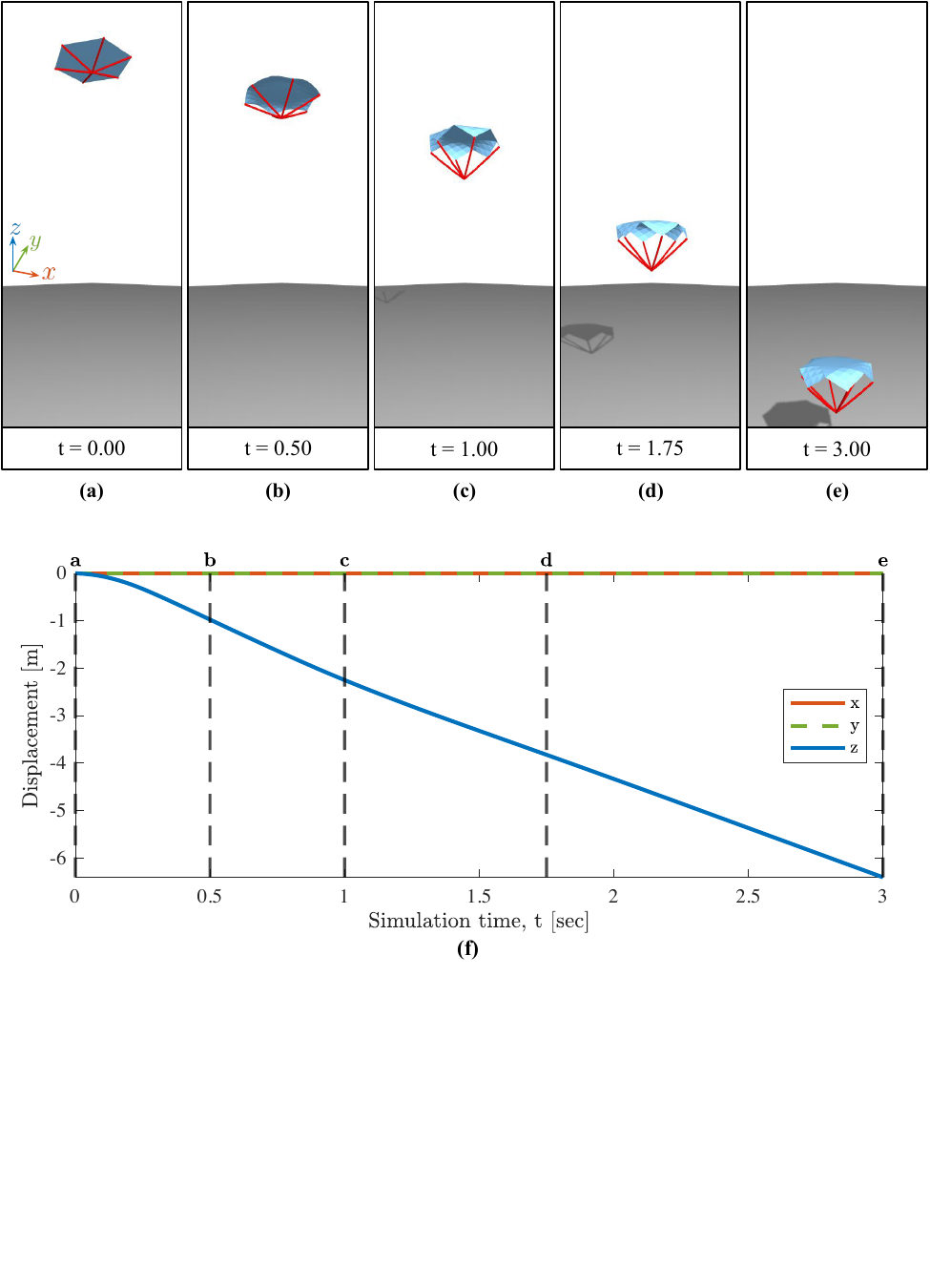}
	\caption{A point mass falling with a parachute modeled as a combination of rods and shell within MAT-DiSMech. (a-e) Snapshots of the parachute at different time stamps. Here the time is given in seconds. (f) Displacement of the hanging mass with time. The time stamps corresponding to the snapshots (a-e) are marked in the plot using vertical dashed lines.} 
\label{fig:rod_shell_robot}
\end{figure}

\textit{Parachute.} 
For our fifth showcase, we simulate a parachute which is modeled as a combination of elastic rods and shell. The canopy is modeled as an initially flat plate in the shape of a regular hexagon of side length $1$ m and thickness $1$ mm. The ropes are modeled as elastic rods. A rigid payload mass of $0.13$ kg is attached at the end of the ropes, which is modeled simply by increasing the mass at the node shared between all rods. The material parameters used in this simulation are: $E_{\textrm{rod}} = 10$ MPa, $ E_{\textrm{shell}} = 1 $ GPa, $ \nu_{\textrm{rod}} = 0.5, \nu_{\textrm{shell}} = 0.3$ and $ \rho = 1500 $ kg/m$^3$ for both rod and shell. The external forces, in this case, are gravity and aerodynamic drag. We use $\mathbf{g} = [0,0,-9.8]^\top$ m/s$^2$, and, $\rho_{\textrm{med}} = 1 $ kg/m$^3$ and $C_D = 10$ to calculate the gravity and drag force respectively. Figure \ref{fig:rod_shell_robot} shows the snapshots of the parachute as it falls to the ground. It can be seen that as the simulation progresses, the initially flat canopy takes a 3D inverted bowl-like shape as expected.

\begin{figure}[h!]
        \centering
	\includegraphics[clip, trim=0cm 4.7cm 0cm 0cm, width=\columnwidth]{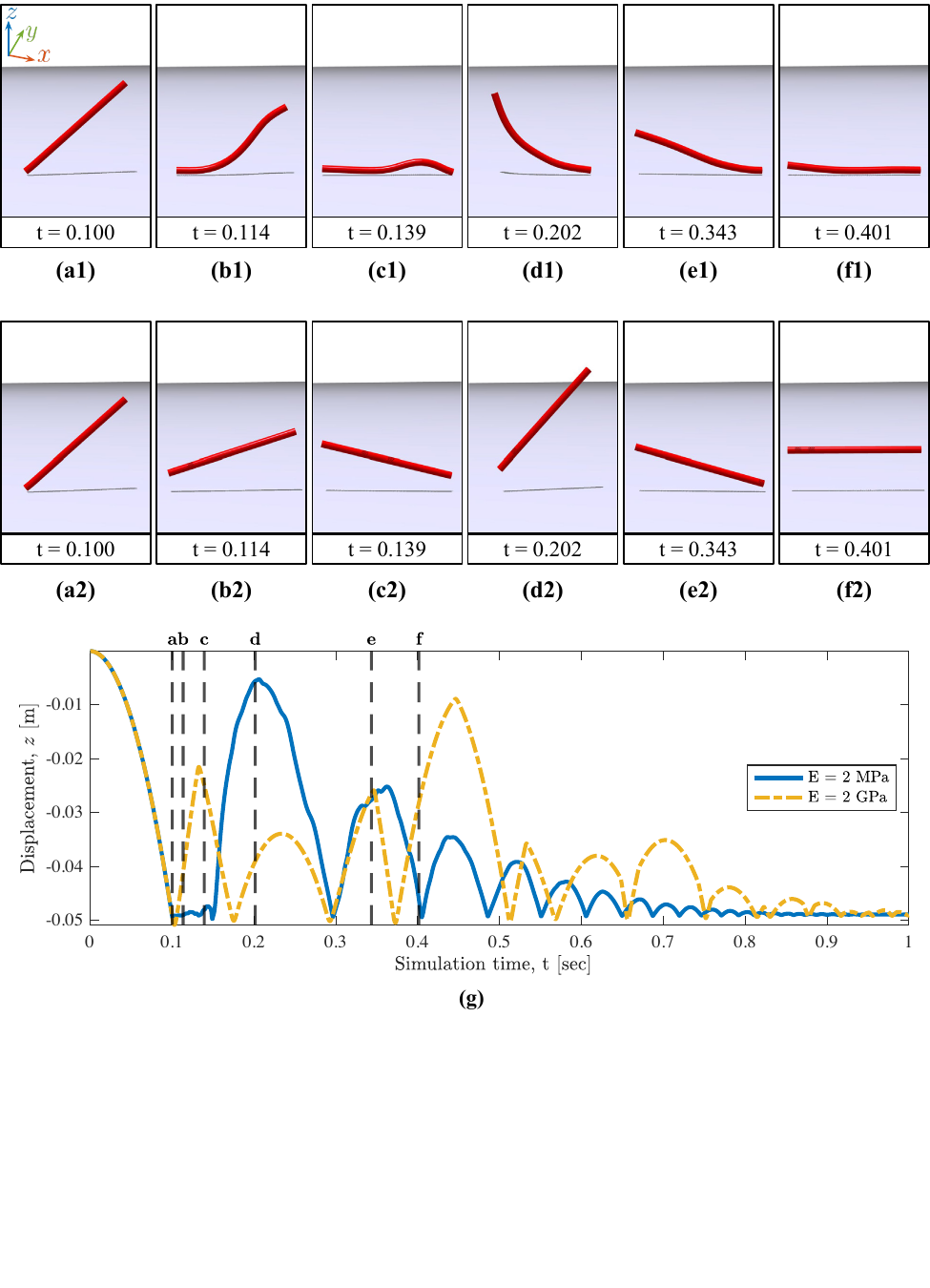}
 \caption{Elastic rod with different Young's Moduli dropped on the ground. (a1-f1) Snapshots of the elastic rod with Young's modulus $2$ MPa, and (a2-f2) Snapshots of the elastic rod with Young's modulus $2$ GPa. Here time is given in seconds. Note that the cross-sectional radius of the rod is $1$ mm, it has been enlarged in the figure for better visualization. (g) The trajectory of the $z$-coordinate of the lowermost point on the rod with time. The time stamps corresponding to the snapshots (a-f) are marked in the plot using vertical dashed lines.} 
\label{fig:rod_contact}
\end{figure}

\textit{Rod dropped on ground.}
For our sixth example, we simulate a rod falling onto the ground from a height of $5$ cm. The density of the rod's material and its Poisson ratio are taken to be $1500$ kg/m$^3$ and $0.3$, respectively. We simulate the rod for two different values of Young's modulus $Y$: (a) $Y = 2$ MPa and (b) $Y = 2$ GPa. The external forces, in this case, are gravity, ground contact, and friction. The values of $[0,0,-9.8]^\top$ m/s$^2$, $0.25$, and $20$ are used respectively for the acceleration due to gravity $\mathbf{g}$, and the friction coefficient $\mu$ and contact stiffness $k_c$ between the ground and the rod. Figure~\ref{fig:rod_contact} shows snapshots of the rod as it hits the ground. As Young's modulus for the second case is much higher than the first case, we observe significant deformation of the rod in the first case while almost no deformation in the second.

\begin{figure}[h!]
    \centering
	\includegraphics[clip, trim=0cm 17.25cm 0cm 0cm, width=\columnwidth]{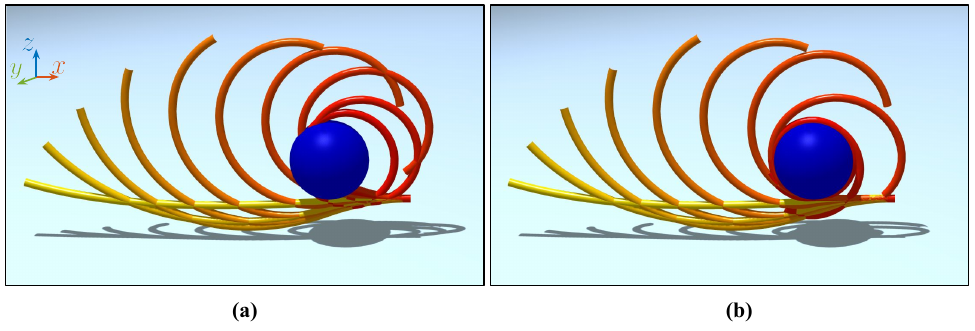}
\caption{Intermediate configurations of a clamped slender rod actuated by varying its natural curvature to bend inward. The rod’s color transitions from yellow to red to indicate progression over time. (a) Without contact modeling, the rod passes through the spherical object to achieve its natural curvature-driven shape. (b) With contact and friction forces modeled via our penalty energy method, the rod interacts with the sphere and successfully wraps around it.}

\label{fig:gripping}
\end{figure}

\textit{Rod gripping a sphere.}
For our final example, we demonstrate a rod gripping a fixed spherical object. The rod has material properties of $E = 200$~MPa, $\nu = 0.5$, and $\rho = 1200$~kg/m$^3$. It measures $0.1$~m in length with a cross-sectional radius of $0.001$~m. Initially positioned along the $x$-axis, the rod is clamped at one end at the origin, while the sphere is located at the point $[0.1, 0.0, 0.05]^\top$. Actuation is performed by gradually varying the natural curvature of the bending-twisting springs along the final three-quarters of the rod’s length, transitioning from $[0, 0]^\top$ to $[0.1886, 0]^\top$. The external forces in this case arise from contact and friction between the rod and the sphere. The simulations use a contact stiffness of 20 and a coefficient of friction of 0.25. Figure~\ref{fig:gripping} illustrates intermediate configurations of a clamped slender rod undergoing actuation via changes in its natural curvature. In case (a), with contact disabled, the rod adopts its natural shape without regard for the spherical obstacle, passing through it if necessary. In contrast, case (b) demonstrates that when contact forces are included through our penalty energy formulation, the rod successfully wraps around the sphere.

In addition to these examples, prior studies have demonstrated that the modeling techniques utilized in our simulator can capture significantly more complex phenomena such as flagella buckling~\cite{PhysRevLettFlagella}, knot tying~\cite{choi_imc_2021}, and shear-induced pitchfork bifurcations in ribbons~\cite{ribbon}. While our examples in this paper are not intended to precisely replicate specific biological trajectories, they are deliberately chosen to showcase the simulator’s ability to model sophisticated motions and handle challenging interactions with the environment -- including contact dynamics, frictional forces, and hydrodynamic resistance.

\subsection{Validation for Cantilever Beam Case}
\begin{figure}[h!]
    \centering
	\includegraphics[clip, trim=0cm 16cm 0cm 0cm, width=\columnwidth]{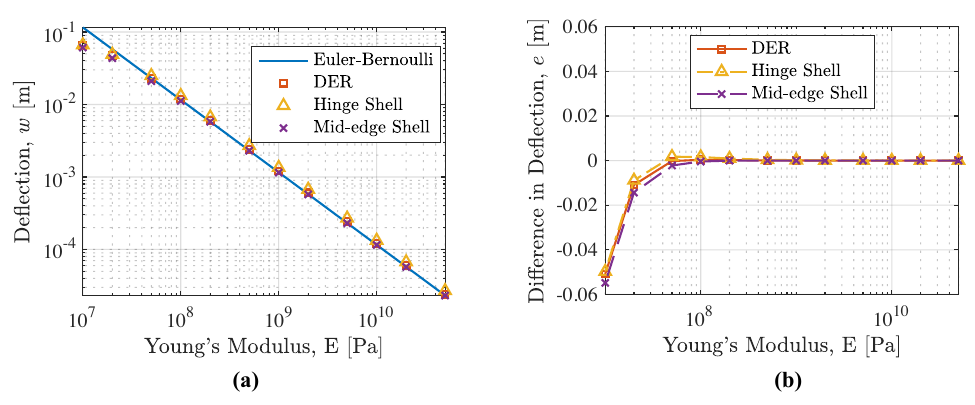}
\caption{Comparison between the cantilever free end deflection obtained from our simulator and the theoretical value from Euler-Bernoulli Beam theory. (a) The log-log plot of the deflection with Young's Modulus for DER, hinge-based and midge-based shell, and Euler Bernoulli beam. (b) A plot of the difference between the value of the deflection obtained from static simulation in MAT-DiSMech and using the Euler-Bernoulli beam theory with Young's modulus in log scale for DER, hinge-based shell and midge-based shell. }
\label{fig:validation}
\end{figure}

In Figure \ref{fig:validation}, we show the plots of the static displacement of the free end of a cantilever beam under gravity. The system is simulated as a rod, a hinge-based shell and mid-edge normal-based shell. The density of the system is $1200$ kg/m$^3$ and the Poisson's ratio is $0.5$. We perform a static simulation in MAT-DiSMech for various Young's moduli, specifically $20$ GPa, $2$ GPa, $200$ MPa and $20$ MPa, and compare the obtained displacement with the theoretical values obtained using the Euler-Bernoulli beam theory \cite{EB}. From Figure~\ref{fig:validation}(a), it is clear that the algorithm is able to capture the deflections in the cantilever beam accurately. Further, from Figure~\ref{fig:validation}(b), it can be seen that the values from the simulator match the values obtained from the Euler-Bernoulli theory more closely at larger values of Young's modulus and move away from the theoretical values at lower values of Young's modulus. This is expected since Euler-Bernoulli's theory is based on approximations which work well for low-deformation cases that correspond to the higher Young's modulus situation.

\begin{figure}[h!]
    \centering
    \includegraphics[clip, trim=0cm 6.2cm 0cm 0cm, width=0.75\columnwidth]{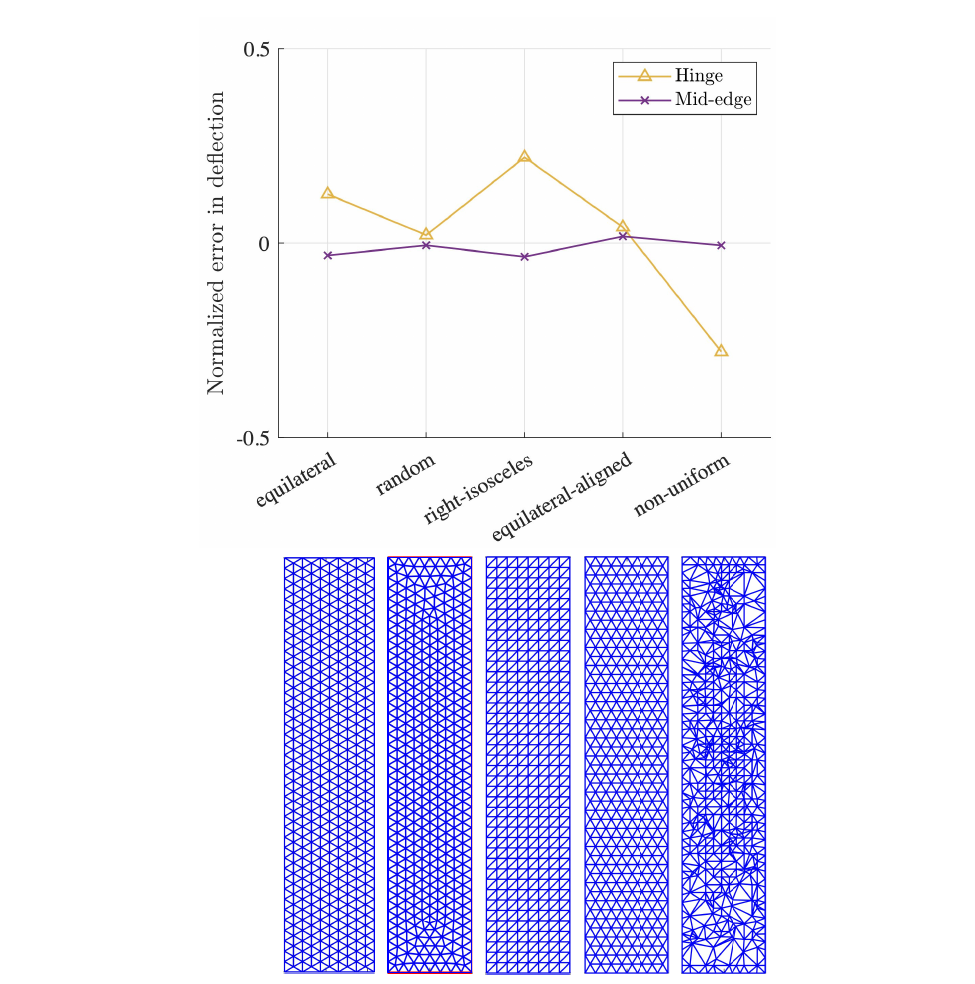}
    \caption{Normalized deflection at the free end of a cantilever beam (length 0.1 m, width 0.02 m, thickness 0.001 m, $E = 2$~GPa, $\nu = 0.5$) under gravity, modeled with hinge and mid-edge bending across different mesh types: equilateral, random, right-isosceles, equilateral-aligned, and non-uniform.}
    \label{fig:mesh_dependence}
\end{figure}

To examine a limitation of the hinge-based bending model for shells, we perform simulations assessing its sensitivity to mesh quality. It is well known that the hinge model performs reliably only on meshes composed of nearly equilateral triangles with good aspect ratios~\cite{grinspun2006}, whereas ill-conditioned meshes with skewed elements or extreme aspect ratios can result in significant errors or convergence issues.
Figure~\ref{fig:mesh_dependence} presents the normalized deflection error for a cantilever beam under gravity, comparing the hinge and mid-edge bending models across five mesh configurations, from highly regular (equilateral) to severely distorted (non-uniform). The results indicate that the mid-edge bending model maintains consistently low errors across all mesh types, demonstrating robustness against mesh irregularities. In contrast, the hinge model shows significant errors for the right-isosceles and non-uniform meshes. The right-isosceles mesh features many hinges oriented obliquely to the main bending direction, while the non-uniform mesh contains triangles with poor aspect ratios and skewness. Overall, these results highlight the hinge method’s mesh dependence, whereas the mid-edge method’s edge-associated degrees of freedom substantially improve robustness to mesh quality.

\section{Conclusions}
The MAT-DiSMech simulation framework presented in this paper offers a comprehensive and versatile approach to simulating soft robotic systems, specifically those composed of rods, shells, or their combinations. By leveraging Discrete Differential Geometry (DDG) methods, our MATLAB-based simulator achieves a balance between accuracy and efficiency, effectively addressing the challenges posed by geometrically nonlinear dynamics in soft robots. The modular design supports a broad range of soft robotic configurations and dynamic forces, including gravity, contact, Coulomb and viscous friction, as well as hydrodynamic and aerodynamic drag, allowing researchers to tailor simulations to specific system requirements. We present a diverse set of showcase examples involving complex environmental interactions, such as frictional ground contact and viscous fluid forces, for both rod- and shell-like structures. We also demonstrate numerical validation against theoretical solutions, highlighting the strengths and limitations of different modeling techniques.

Since controlled actuation plays a central role in soft robotics, MAT-DiSMech supports time-dependent adjustments of natural stretch, curvature, and twist. This enables the simulation of dynamic behaviors such as shape morphing, as demonstrated in examples including an earthworm, manta ray, snake, PneuNet actuator, and gripping manipulator. In conclusion, MAT-DiSMech provides an accurate and efficient platform for simulating soft robot dynamics. It facilitates rapid prototyping and validation of diverse soft robotic designs, thereby advancing research along Sim2Real pathways. In the future, we plan on expanding the scope of MAT-DiSMech by integrating data-driven modeling techniques to model material nonlinearity and black-box external forces.


\section*{Acknowledgements} 
We acknowledge financial support from the National Science Foundation (grant numbers 20476630, 2209782) and the National Institute of Neurological Disorders and Stroke (1R01NS141171-01).




\section*{Appendix}
\label{sec:appendix}
\begin{appendices}
\section{Mid-edge Normal Bending}
\label{AppendixA:Midedge}

As explained in Section \ref{sec: midedge}, the per-triangle bending energy density for a shell in terms of the shape operator $\Lambda$ is given by
\begin{equation*}
    E^b = \frac{Yh^3}{1-\nu^2}[(1-\nu)\mathrm{Tr}((\Lambda-\bar{\Lambda})^2) +  \nu(\mathrm{Tr}(\Lambda) - \mathrm{Tr}(\bar{\Lambda}) )^2],
\end{equation*}
where
\begin{align*}
    &\Lambda = \sum_{k=1}^{3} c^k(s^k\xi^k - f^k) \mathbf{t}^k  \otimes \mathbf{t}^k, \quad \textrm{and} \\
    &\bar{\Lambda} = \sum_{k=1}^{3} \bar{c}^k(s^k\bar{\xi}^k - \bar{f}^k) \mathbf{\bar{t}}^k  \otimes \mathbf{\bar{t}}^k.
\end{align*}

Here $c^k = \frac{1}{\bar{A}_i \|\mathbf{\bar{e}}^k\| (\hat{\mathbf{t}}^k \cdot \boldsymbol{\tau}^{k,0})} $ and $f^k = \mathbf{n}\cdot{\boldsymbol{{\tau}}^{k,0}}$ 
and $\bar{}$ denotes the values at the undeformed or natural configuration.
Note that we have omitted the subscript $i$, denoting the $i$-th triangle for brevity. 
To find the bending force and jacobian, we compute the gradient and hessian of the bending energy with respect to the DOF vector. We use the following simplification as suggested in the work of Grinspun et al. \cite{grinspun2006}. For thin shells, the membrane term dominates the bending term (since the bending stiffness is proportional to the third power of thickness ($h^3$) and the membrane term is proportional to $h$) and the shape of the triangular faces is mostly preserved. In such a case, only the derivative of  $f_i$'s with respect to $\mathbf{p}_i$'s are needed, and the derivatives of $c_i$'s and $t_i$'s with respect to $\mathbf{p}_i$'s can be neglected.

Let $E_I = \mathrm{Tr}((\Lambda-\bar{\Lambda})^2)$ and $E_{II} = (\mathrm{Tr}(\Lambda) - \mathrm{Tr}(\bar{\Lambda}) )^2$. The expressions for the gradient and the hessian of these terms are provided below.

\noindent
\underline{Gradient} $(12\times 1)$:
\begin{align*}
    \frac{\partial E_I}{\partial \mathbf{q}} &= \frac{\partial \mathrm{Tr}((\Lambda-\bar{\Lambda})^2)}{\partial \mathbf{q}} = \frac{\partial \mathrm{Tr}(\Lambda^2)}{\partial \mathbf{q}} - 2\frac{\partial(\mathrm{Tr}(\Lambda \bar{\Lambda}))}{\partial \mathbf{q}},
\end{align*}
where the second equality above is due to the fact that $\bar{\Lambda}$ is a constant.
Then, since the DOFs for the mid-edge normal method include $\mathbf{x}_k$ and $\xi^i$, where $i,k \in \{1,2,3\}$ as explained in Section \ref{sec: midedge}, we solve for the derivatives of the above two terms with respect to these DOFs as follows:
\begin{align*}
    \frac{\partial \mathrm{Tr}(\Lambda^2)}{\partial \mathbf{x}_k} &= \sum_{i=1}^3\sum_{j=1}^3 - c^i c^j \left [(s^i \xi^i - f^i)\frac{\partial f^j}{\partial \mathbf{x}_k} + \right. 
    \\
    &\left. \quad \quad \quad \quad \quad \quad (s^j \xi^j - f^j) \frac{\partial f^i}{\partial \mathbf{x}_k}\right ] (\mathbf{t}^i\cdot \mathbf{t}^j)^2,
    \\
    \frac{\partial(\mathrm{Tr}(\Lambda \bar{\Lambda}))}{\partial \mathbf{x}_k} &= \sum_{i=1}^3 \sum_{j=1}^3 -c^i \bar{c}^j (s^j\bar{\xi}^j -\bar{f}^j) \frac{\partial f^i}{\partial \mathbf{x}_k} (\mathbf{t}^i\cdot \mathbf{\bar{t}}^j)^2,
    \\
    \frac{\partial \mathrm{Tr}(\Lambda^2)}{\partial \xi^i} &= 2 \hspace{1pt} c^i \hspace{1pt} s^i \hspace{1pt} \sum_{j=1}^3 c^j (s^j\xi^j -f^j) (\mathbf{t}^i\cdot \mathbf{t}^j)^2,
    \\
    \frac{\partial (\mathrm{Tr}(\Lambda \bar{\Lambda}))}{\partial \xi^i} &= c^i s^i \sum_{j=1}^3 \bar{c}^j (s^j\bar{\xi}^j -\bar{f}^j) (\mathbf{t}^i\cdot \mathbf{\bar{t}}^j)^2.
\end{align*}

Performing similar calculations for $E_{II}$, we obtain the following:
\begin{align}
\label{eq:intermediate}
    \frac{\partial E_{II}}{\partial \mathbf{q}} = \frac{\partial (\mathrm{Tr}(\Lambda))^2}{\partial \mathbf{q}} - 2\mathrm{Tr}(\bar{\Lambda})\frac{\partial(\mathrm{Tr}(\Lambda))}{\partial \mathbf{q}}.
\end{align}
Then,
\begin{align*}
    \frac{\partial (\mathrm{Tr}(\Lambda))^2}{\partial \mathbf{x}_k} &= \sum_{i=1}^3\sum_{j=1}^3 - c^i c^j {l^i}^2 {l^j}^2\left [(s^i \xi^i - f^i)\frac{\partial f^j}{\partial \mathbf{x}_k}  \right. 
    \\
    &\left. \qquad \qquad \qquad \qquad  \quad +(s^j \xi^j - f^j) \frac{\partial f^i}{\partial \mathbf{x}_k}\right ].
\end{align*}
    Furthermore, the computation of the second term in \eqref{eq:intermediate} results in the following:
\begin{align*}
    \mathrm{Tr}(\bar{\Lambda})\frac{\partial \mathrm{Tr}(\Lambda)}{\partial \mathbf{x}_k} &= \sum_{i=1}^3 \sum_{j=1}^3 -c^i {\bar{c}}^j {{l^i}^2} ({\bar{l}^j)^2} (s^j{\bar{\xi}}^j-{\bar{f}}^j)\frac{\partial f^i}{\partial \mathbf{x}_k},
    \\
    \frac{\partial (\mathrm{Tr}(\Lambda))^2}{\partial \xi^i} &= 2 c^i s^i {l^i}^2 \sum_{j=1}^3 c^j {l^j}^2 (s^j\xi^j -f^j),
    \\
    \mathrm{Tr}(\bar{\Lambda})\frac{\partial \mathrm{Tr}(\Lambda)}{\partial \xi^i} &= c^i s^i {l^i}^2 \sum_{j=1}^3 \bar{c}^j (\bar{l}^j)^2(s^j\bar{\xi}^j -\bar{f}^j),
\end{align*}
where $\frac{\partial f^i}{\partial \mathbf{x}_{k}} = \frac{(\tau^{i,0} \cdot \mathbf{t}^k) \mathbf{n}}{2A}$.\\

Next, the computation of the Hessian of $E_I$ and $E_{II}$ is as follows. For brevity, we write $\mathbf{q} = [q_1, q_2,\ldots,q_{12}]^\top$. 

\underline{Hessian} $(12\times 12)$:
\begin{align*}
    \frac{\partial^2 E_I}{\partial q_i \partial q_j} = \frac{\partial^2 \mathrm{Tr}(\Lambda^2)}{\partial q_i \partial q_j} - 2\frac{\partial^2(\mathrm{Tr}(\Lambda \bar{\Lambda}))}{\partial q_i \partial q_j}.
\end{align*}
Then, we compute each of the terms in the above expression of the Hessian as follows:
\begin{align*}
    \frac{\partial^2 (\mathrm{Tr}(\Lambda^2))}{\partial \mathbf{x}_{k1} \partial \mathbf{x}_{k2}} &= \sum_{i=1}^3 \sum_{j=1}^3 \hspace{1pt} c^i c^j  \left[(s^i \xi^i - f^i)\frac{\partial^2 f^j}{\partial \mathbf{x}_{k1} \partial \mathbf{x}_{k2}}\right.
    \\
   \qquad \qquad \:\: &\left.-\left (\frac{\partial f^i}{\partial \mathbf{x}_{k2}}\right )^\top \frac{\partial f^j}{\partial \mathbf{x}_{k1}} - \left (\frac{\partial f^j}{\partial \mathbf{x}_{k2}}\right )^\top \frac{\partial f^i}{\partial \mathbf{x}_{k1}}\right.
   \\
 \qquad \qquad \:\: &\left. + (s^j \xi^j - f^j) \frac{\partial^2 f^i}{\partial \mathbf{x}_{k1} \partial \mathbf{x}_{k2}} \right ] (\mathbf{t}^{i}\cdot \mathbf{t}^{j})^2,
 \\
 \frac{\partial^2 (\mathrm{Tr}(\Lambda^2))}{\partial \xi^{i} \partial \xi^{j}} &= 2 \hspace{1pt} c^{i} \hspace{1pt} c^{j} \hspace{1pt}  s^{i} \hspace{1pt} s^{j} \hspace{1pt} (\mathbf{t}^{i}\cdot \mathbf{t}^{j})^2 ,
 \\
   \frac{\partial^2 (\mathrm{Tr}(\Lambda^2))}{\partial \xi^{i} \partial \mathbf{x}_{k}} &= -2 \hspace{1pt} c^{i} \hspace{1pt} s^{i} \hspace{1pt} \sum_{j=1}^3 \hspace{1pt} c^j \frac{\partial f^j}{\partial \mathbf{x}_{k}} (\mathbf{t}^{i}\cdot \mathbf{t}^j)^2,
   \\
    \frac{\partial^2(\mathrm{Tr}(\Lambda \bar{\Lambda}))}{\partial \mathbf{x}_{k1} \partial \mathbf{x}_{k2}} &= \sum_{i=1}^3 \sum_{j=1}^3 \hspace{1pt} c^i \bar{c}^j \bar{f}^j \frac{\partial^2 f^i}{\partial \mathbf{x}_{k1} \partial \mathbf{x}_{k2}} (\mathbf{t}^i\cdot \mathbf{\bar{t}}^j)^2 ,
    \\
    \frac{\partial^2(\mathrm{Tr}(\Lambda \bar{\Lambda}))}{\partial \xi^i \partial \xi^j} &= 0 ,\quad \quad
     \frac{\partial^2(\mathrm{Tr}(\Lambda \bar{\Lambda}))}{\partial \mathbf{x}_k \partial \xi^i} = 0.
\end{align*}

\noindent
Similarly, for $E_{II}$, computing the Hessian results in the following:
\begin{align*}
    \frac{\partial^2 E_{II}}{\partial q_i \partial q_j} = \frac{\partial^2 (\mathrm{Tr}(\Lambda))^2}{\partial q_i \partial q_j} - 2\mathrm{Tr}(\bar{\Lambda})\frac{\partial^2(\mathrm{Tr}(\Lambda ))}{\partial q_i \partial q_j}. 
\end{align*}
Then,
\begin{align*}
    \frac{\partial^2 (\mathrm{Tr}(\Lambda))^2}{\partial \mathbf{x}_{k1} \partial \mathbf{x}_{k2}} &= \sum_{i=1}^3 \sum_{j=1}^3 \hspace{1pt} c^i c^j \left[(s^i \xi^i - f^i)\frac{\partial^2 f^j}{\partial \mathbf{x}_{k1} \partial \mathbf{x}_{k2}}\right.
    \\
   \qquad \qquad \;\; &\left.-\left (\frac{\partial f^i}{\partial \mathbf{x}_{k2}}\right )^\top \frac{\partial f^j}{\partial \mathbf{x}_{k1}} - \left (\frac{\partial f^j}{\partial \mathbf{x}_{k2}}\right )^\top \frac{\partial f^i}{\partial \mathbf{x}_{k1}}\right.
   \\
 \qquad \qquad \;\; &\left. + (s^j \xi^j - f^j) \frac{\partial^2 f^i}{\partial \mathbf{x}_{k1} \partial \mathbf{x}_{k2}} \right ]{l^i}^2 {l^j}^2,
 \\
 \frac{\partial^2 (\mathrm{Tr}(\Lambda))^2}{\partial \xi^{i} \partial \xi^{j}} &= 2 \hspace{1pt} c^{i} \hspace{1pt} c^{j} \hspace{1pt}  s^{i} \hspace{1pt} s^{j} \hspace{1pt} {l^i}^2 {l^j}^2,
 \\
 \frac{\partial^2 \mathrm{Tr}(\Lambda)}{\partial \mathbf{x}_{k1} \partial \mathbf{x}_{k2}} &= -\sum_{i=1}^3 c^i {l^i}^2 \frac{\partial^2 f^i}{\partial \mathbf{x}_{k1} \partial \mathbf{x}_{k2}},
 \\
   \frac{\partial^2 (\mathrm{Tr}(\Lambda))^2}{\partial \xi^{i} \partial \mathbf{x}_{k}} &= -2 \hspace{1pt} c^{i} \hspace{1pt} s^{i} {l^i}^2 \hspace{1pt} \sum_{j=1}^3 \hspace{1pt} c^j {l^j}^2 \frac{\partial f^j}{\partial \mathbf{x}_{k}} ,
    \\
    \frac{\partial^2(\mathrm{Tr}(\Lambda))}{\partial \xi^i \partial \xi^j} &= 0, \quad \quad
     \frac{\partial^2(\mathrm{Tr}(\Lambda ))}{\partial \mathbf{x}_k \partial \xi^i} = 0,
\end{align*}
\noindent
where \[\frac{\partial^2 f^i}{\partial \mathbf{x}_{k1} \partial \mathbf{x}_{k2}} = \frac{1}{4A^2}\left( (\tau^{i,0} \cdot \mathbf{t}^{k1}) \: (\mathbf{n} \otimes \mathbf{t}^{k2} + \mathbf{t}^{k2} \otimes \mathbf{n} ) \right )\]
where the operator $\otimes$~denotes the outer product.

\section{RFT Jacobian Expressions}
\label{AppendixB:Visc}
For the Jacobian of the RFT hydrodynamic force, we provide the expressions for 
$\frac{\partial \hat{\mathbf{e}}^k}{\partial \mathbf{x}_j}$ and $\frac{\partial \mathbf{u}_i}{\partial \mathbf{x}_j}$ used in equation \eqref{eq:JRFT}.

\begin{equation}
\label{eq:RFT_appendix}
    \frac{\partial \hat{\mathbf{e}}^k}{\partial \mathbf{x}_j} = 
    \begin{cases}
        \frac{\mathbf{e}^k{\mathbf{e}^k}^\top - ||\mathbf{e}^k||^2\mathbb{I}}{||\mathbf{e}^k||^3} & \mathbf{e}^k=\mathbf{x}_n - \mathbf{x}_j,
        \\
        \frac{||\mathbf{e}^k||^2\mathbb{I} - \mathbf{e}^k{\mathbf{e}^k}^\top}{||\mathbf{e}^k||^3} & \mathbf{e}^k=\mathbf{x}_j - \mathbf{x}_n,
        \\
        0  & \textrm{otherwise}
    \end{cases}
\end{equation}
where $n$ can be any integer and $\mathbb I$ is the square identity matrix of size $(3\times 3)$.

\begin{equation}
\label{eq:RFT_appendix2}
    \frac{\partial \mathbf{u}_i}{\partial \mathbf{x}_j} = 
    \begin{cases}
        \frac{\mathbb{I}}{dt} & i=j, 
        \\
        0  & i \neq j 
    \end{cases}
\end{equation}
where $\mathbb I$ is the square identity matrix of size $(3\times 3)$.

\section{Aerodynamic Drag Jacobian Expressions}
\label{AppendixC:Aero}
For the Jacobian of the aerodynamic drag force, we provide the expression for 
$\frac{\partial \hat{\mathbf{n}}_k}{\partial \mathbf{x}_j}$ and $\frac{\partial \mathbf{u}_i}{\partial \mathbf{x}_j}$ used in equation \eqref{eq: Jdrag}. If $t$-th triangle has the nodes $\mathbf{x}_i, \mathbf{x}_j, \mathbf{x}_k$, then,
\begin{equation}
    \frac{\partial \mathbf{n}_t}{\partial \mathbf{x}_i} = 
        \frac{1}{||\mathbf{n}_t||^3}\left[||\mathbf{n}_t||^2\mathbb{I} - \mathbf{n}_t\mathbf{n}_t^\top\right][(\mathbf{x}_k - \mathbf{x}_j)\times]
\end{equation}
where $\mathbb I$ is the square identity matrix of size $(3\times 3)$ and $[\mathbf{p}\times]$ denotes the $(3\times 3)$ skew-symmetric cross product matrix for a vector $\mathbf{p}$. $\frac{\partial \mathbf{n}_t}{\partial \mathbf{x}_n} = 0$ if $n\not\in\{i,j,k\}$.

The expression for $\frac{\partial \mathbf{u}_i}{\partial \mathbf{x}_j}$ stays the same as in \eqref{eq:RFT_appendix2}.

\end{appendices}

\end{multicols}
\end{document}